\documentclass[pdflatex,sn-mathphys-num]{sn-jnl}

\usepackage{graphicx}%
\usepackage{multirow}%
\usepackage{amsmath,amssymb,amsfonts}%
\usepackage{amsthm}%
\usepackage{mathrsfs}%
\usepackage[title]{appendix}%
\usepackage{xcolor}%
\usepackage{textcomp}%
\usepackage{manyfoot}%
\usepackage{booktabs}%
\usepackage{algorithm}%
\usepackage{algorithmicx}%
\usepackage{algpseudocode}%
\usepackage{listings}%

\usepackage[utf8]{inputenc} 
\usepackage[T1]{fontenc}    

\usepackage{xr}
\usepackage{cleveref}
\usepackage{hyperref}       
\usepackage{url}            
\usepackage{booktabs}       
\usepackage{amsfonts}       
\usepackage{nicefrac}       
\usepackage{microtype}      
\usepackage{lipsum}		
\usepackage{graphicx}
\usepackage{natbib}
\setcitestyle{super}
\usepackage{doi}
\usepackage{enumitem}
\usepackage{subfigure}
\usepackage{verbatim}
\usepackage{makecell}
\usepackage[strings]{underscore}

\usepackage{xspace}  

\usepackage{hyperref}
\hypersetup{hidelinks,
	colorlinks=true,
	pdfstartview=Fit,
	breaklinks=true}

\usepackage{url}
\usepackage{bm}  
\usepackage{amsmath}  
\usepackage{upgreek}  

\theoremstyle{thmstyleone}%
\theoremstyle{thmstyletwo}%

\theoremstyle{thmstylethree}%

\newcommand{\thename}[0]{JCDI\xspace}
\colorlet{cyan}{.}

\begin{document}

\title[Article Title]{Diffusion Model-based Parameter Estimation in Dynamic Power Systems}


\author[1,4]{\fnm{Feiqin} \sur{Zhu}}\email{fqzhu@suda.edu.cn}  
\equalcont{These authors contributed equally.}

\author[2]{\fnm{Dmitrii} \sur{Torbunov}}\email{dtorbunov@bnl.gov}
\equalcont{These authors contributed equally.}

\author[3,5]{\fnm{Zhongjing} \sur{Jiang}}\email{zjiang35@illinois.edu}

\author[1,6]{\fnm{Tianqiao} \sur{Zhao}}\email{tianqiao.zhao@uta.edu}

\author[1]{\fnm{Amirthagunaraj} \sur{Yogarathnam}}\email{rajyogar@ieee.org}

\author*[2]{\fnm{Yihui} \sur{Ren}}\email{yren@bnl.gov}  

\author[1]{\fnm{Meng} \sur{Yue}}\email{yuemeng@bnl.gov}

\affil[1]{\orgdiv{Interdisciplinary Science Department}, \orgname{Brookhaven National Laboratory}, \orgaddress{\city{Upton}, \postcode{11973-5000}, \state{NY}, \country{USA}}}


\affil[2]{\orgdiv{Computing and Data Sciences Directorate}, \orgname{Brookhaven National Laboratory}, \orgaddress{\city{Upton}, \postcode{11973-5000}, \state{NY}, \country{USA}}}


\affil[3]{\orgdiv{Environmental Science and Technologies Department}, \orgname{Brookhaven National Laboratory}, \orgaddress{ \city{Upton}, \postcode{11973-5000}, \state{NY}, \country{USA}}}

\affil[4]{Present address: \orgdiv{School of Rail Transportation}, \orgname{Soochow University}, \orgaddress{ \city{Suzhou}, \postcode{215131}, \country{China}}}

\affil[5]{Present address: \orgdiv{Institute for Sustainability, Energy, and Environment}, \orgname{University of Illinois Urbana-Champaign}, \orgaddress{ \city{Urbana}, \postcode{61801}, \state{IL}, \country{USA}}}

\affil[6]{Present address: \orgdiv{Department of Electrical Engineering}, \orgname{University of Texas at Arlington}, \orgaddress{ \city{Arlington}, \postcode{76019}, \state{TX}, \country{USA}}}


\abstract{
Parameter estimation, which represents a classical inverse problem, is often ill-posed as different parameter combinations can yield identical outputs. This non-uniqueness presents a critical barrier to accurate and unique identification. 
\textcolor {cyan}{Here we introduce} a parameter estimation framework to address such limits: the Joint Conditional Diffusion Model-based Inverse Problem Solver.
By \textcolor {cyan}{leveraging} the stochasticity of diffusion models, \textcolor {cyan}{it} produces candidate solutions that \textcolor {cyan}{capture underlying parameter distributions conditioned on the observations}. 
Joint conditioning on multiple observations further narrows the posterior distributions of non-identifiable parameters.
For composite load model parameterization, a challenging task in dynamic power systems, \textcolor {cyan}{the proposed method} achieves a 58.6\% reduction in parameter estimation error compared to the single-condition model.
It also accurately replicates system's dynamic responses under various electrical faults with root mean square errors below $4 \times {10^{ - 3}}$, 
exhibiting comprehensive advantages in calibration and efficiency over existing methods.
Given its data-driven nature, \textcolor {cyan}{it} provides a \textcolor {cyan}{general} framework for parameter estimation while effectively mitigating the non-uniqueness problem across scientific domains.
}




\maketitle

\section*{Introduction}
\label{sec1}

Dynamical system modeling is fundamental across many scientific disciplines, from synoptic meteorology to electric power grids.
While forward modeling, which predicts system evolution from initial states, has received extensive attention, 
many practical problems in system dynamics, such as parameter identification, are framed as inverse problems, where inherent properties are deduced from observed behaviors~\citep{book1-inverse}. 
For systems with complex dynamics, such inference becomes particularly challenging due to the non-uniqueness of solutions, which is also \textcolor {cyan}{a manifestation of} parameter non-identifiability, where different parameter combinations can produce identical system behaviors~\citep{book1-inverse,Lederman2021}.

The modern power system presents a compelling case for studying the inverse problem in system dynamics.  
Electric power grids are evolving with the integration of emerging and flexible loads, including data centers, electrified transportation, and buildings, as well as the widespread adoption of power electronic devices~\citep{World-Energy,van2025demand,powell2022charging}. 
This evolution has dramatically increased the complexity and uncertainty of power system dynamics.
Electrical loads, constituting a fundamental component of power systems, are characterized by their numerous quantities, spatial dispersion, and heterogeneity~\citep{jacob2024real}. Accurate load modeling is crucial for analyzing contingencies and stability in large-scale power systems while presenting unique challenges~\citep{Dynamic-Performance}.
Load aggregation addresses issues with the aggregating numerous end user load devices by consolidating various individual power loads into an equivalent model, reducing computational complexity yet preserving essential behavioral characteristics.
The composite load model (CLM) represents a class of component-based aggregated models that categorizes individual end user load devices into several component groups based on their electrical characteristics~\citep{load-review}.
Nevertheless, the CLM operates as a gray box model with specified structure but unknown parameters.

\begin{figure}[htbp]
	\centering
    \includegraphics[width=0.9\textwidth]{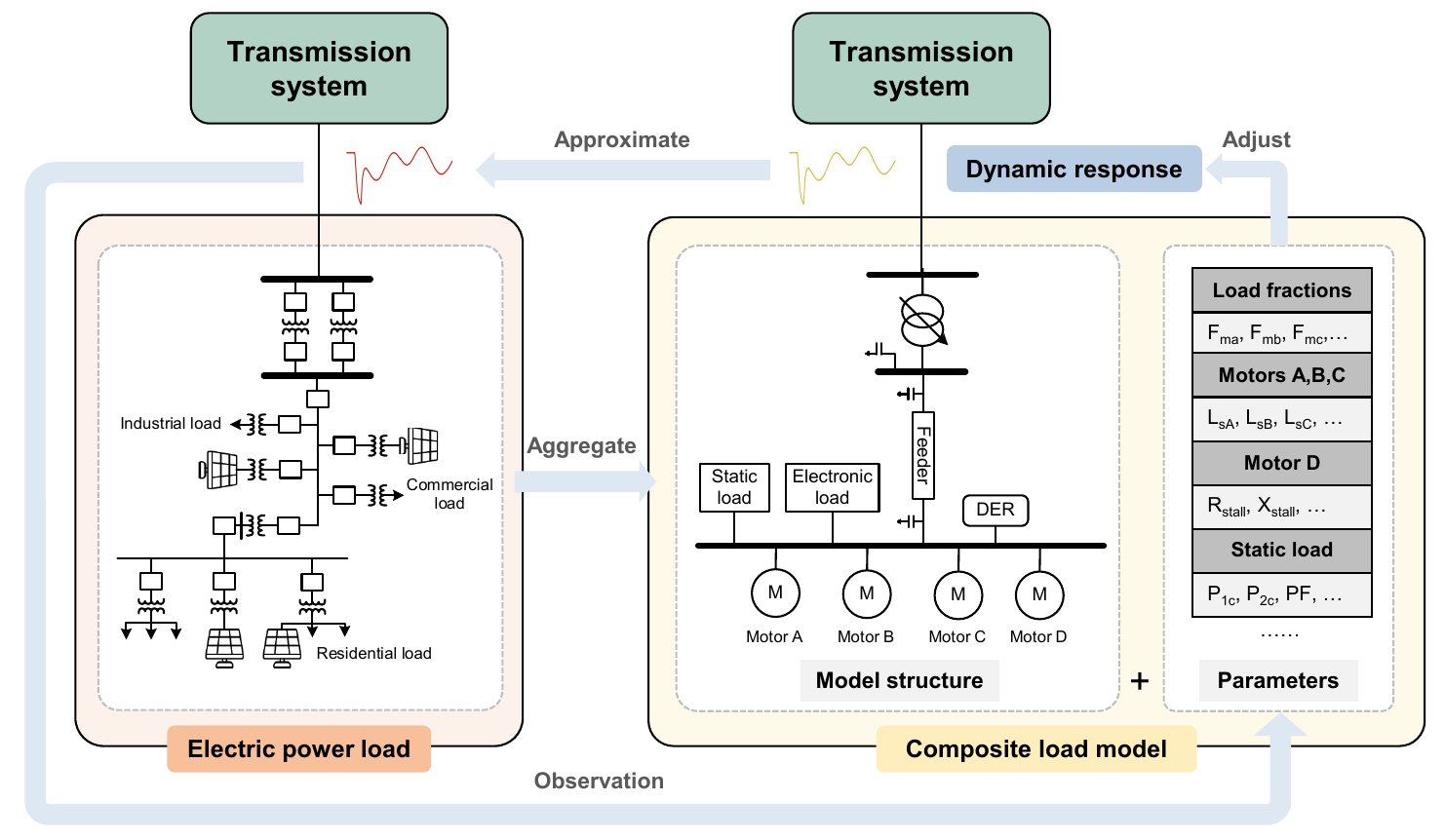} 
	\caption{\textbf{Parameterization for the composite load model.} CLM uses several types of load devices to represent the aggregation behavior of the numerous power loads. The parameterization of CLM represents a classical inverse problem, which deduces model parameters from the observed dynamic responses. With the estimated parameters, CLM is expected to approximate its dynamic behavior to real measurements.}
	\label{fig:WECC-CLM}
\end{figure}

The proliferation of grid sensing devices, such as phasor measurement units, provides extensive measurement data that capture the dynamic behaviors of electric loads under various disturbances. 
The parameterization of CLM from these measurements exemplifies a classical inverse problem (Fig.~\ref{fig:WECC-CLM}).
Despite \textcolor {cyan}{considerable} simplifications from detailed load representations, CLM still encompasses dozens of differential algebraic equations with almost 200 parameters to be determined. 
The inverse problem of parameterization for such a complex, nonlinear dynamical system typically is ill-posed, meaning the solution may not be unique, and minor input variations can produce dramatic changes of output, yielding unstable results~\citep{ASTER201955,piani2024data,engl2014inverse}.
This non-uniqueness of parameter estimates limits generalization across different dynamic disturbances, i.e., CLM parameters carefully selected under one electrical fault-induced disturbance may exhibit inadequate performance in another fault. 

Data-driven machine learning has emerged as a promising paradigm for modeling complex system dynamics~\citep{wang2023scientific,hamann2024foundation,aarts2025physics}.
Recent advances, such as the physics-informed neural network and conditional generative adversarial networks, have demonstrated they can provide more accurate and efficient solutions for both forward and inverse problems~\citep{raissi2019physics,kadeethum2021framework,gabbard2022bayesian}. 
In the context of CLM parameterization, reinforcement learning (RL) has been extensively investigated~\citep{Two-Stage,Evolutionary-RL,Imitation-Q}. It reformulates parameter calibration as a Markov decision process, where policies are learned to adjust model parameters toward minimal observation error through trial and error. 
However, it tends to fall into local optimum, especially when the action space is large.
Additionally, RL requires iterative reevaluation with forward simulation for each trial of model parameters, which becomes computationally expensive when the grid simulation model is complex.
Supervised learning (SL) neural networks, such as the multi-residual deep \textcolor {cyan}{neural network~\citep{Multi-Residual}}, present an alternative approach. They extract the observation features and directly learn the mapping between observations and model parameters. 
However, these networks often fail to represent the one-to-many mapping between observations and parameters~\citep{Toward-Online}.  
Recent advances in probabilistic approaches include deep generative architectures for time-varying parameter identification~\citep{Probabilistic-Generative} and amortized Bayesian estimation using conditional masked autoregressive flow~\citep{CMAF}.
While these methods effectively manage uncertainty, they do not address the parameter non-uniqueness problem. 
Recent RL investigations have explored multi-event training and multi-task learning strategies to cope with this challenge~\citep{Toward-Online,Imitation-Q}. 
However, non-stationarity in multi-event environments can degrade the performance of learning agents, and multi-task learning risks negative transfer between tasks.

As an emerging family of deep generative models, diffusion models construct synthetic data samples by gradually adding random noise to data then learning to reverse the diffusion process~\citep{DDPM}. 
The model faithfully learns the probability distribution of a dataset and provides a means to generate samples from this distribution.
It has achieved state-of-the-art generation quality compared with existing generative models, such as generative adversarial networks~\citep{diffusion-survey,dhariwal2021diffusion,du2023beyond,price2025probabilistic}. 
Diffusion models also can be trained in a conditioned form, allowing flexible control over the generation process toward the expected style. 
Along with forward generation, the denoising technology represents a compelling approach for solving inverse problems by implicitly encoding valuable information from input signals~\citep{daras2024surveydiffusionmodelsinverse}.
Diffusion-model-based solutions have been successfully developed for a variety of inverse problems, including image restoration and super-resolution~\citep{kawar2022denoising}. However, \textcolor {cyan}{few existing studies involve} the inverse problem applied in dynamical systems like parameter estimation.

\begin{figure}[htbp]
	\centering
    \includegraphics[width=0.9\textwidth]{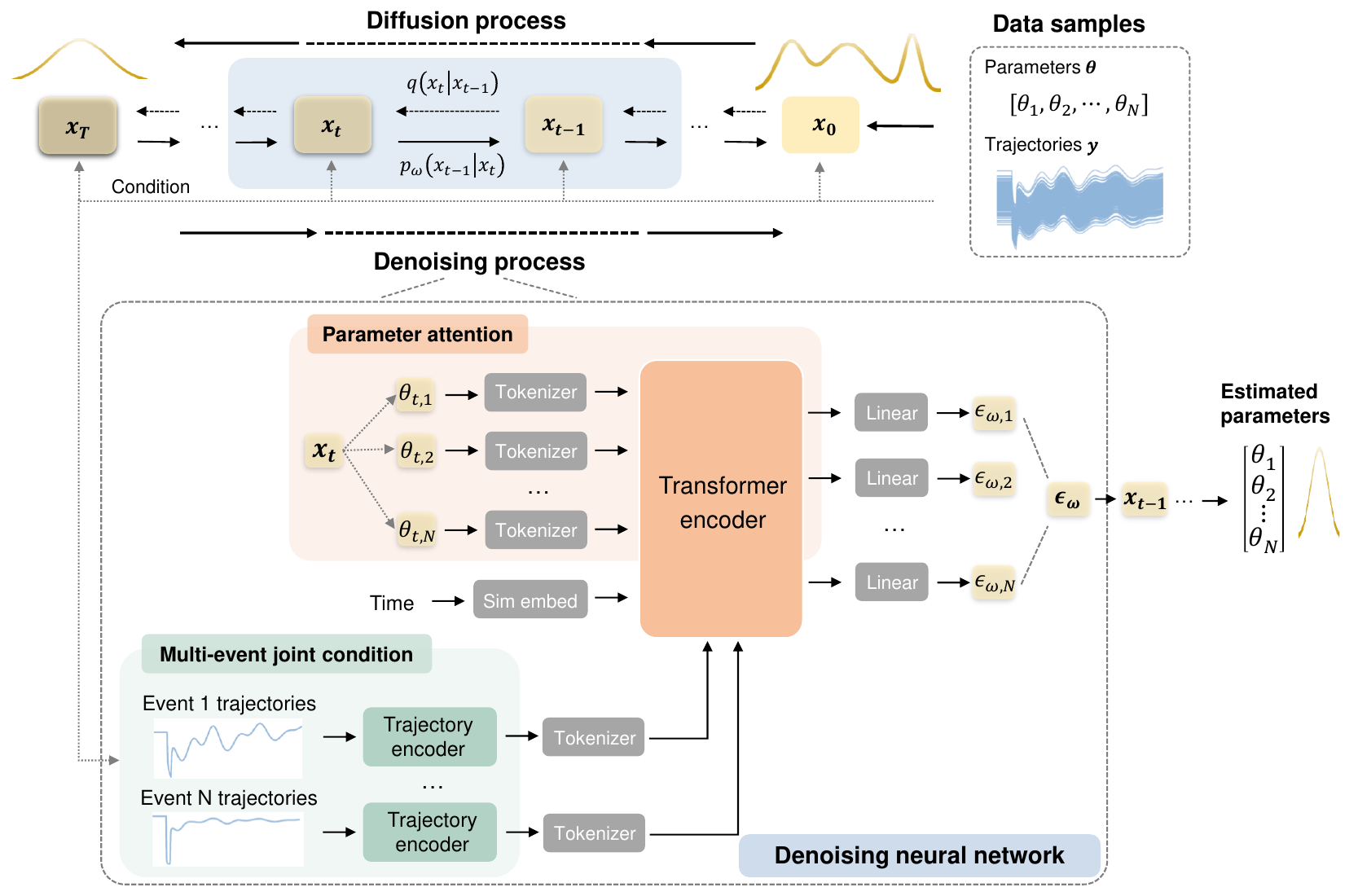} 
	\caption{\textbf{Illustration of JCDI Framework.} This framework is based on a conditional diffusion model. 
    Diffusion and reverse processes are applied on the model parameter space.
    Systems observations serve as the conditions to guide the generation process.
    The denoising neural network is constructed based on a transformer encoder. 
    Three types of inputs -- (1) model parameters \textcolor {cyan}{with added noise}, (2) encoded power trajectories, and (3) encoded diffusion time step -- are tokenized and fed into the neural network.
    The multi-event joint conditioning mechanism is employed, allowing the diffusion model to be simultaneously conditioned on multiple observations under various dynamic disturbances.}
	\label{fig:JCDI}
\end{figure}

Depicted in Figure~\ref{fig:JCDI}, we present a probabilistic parameter estimation framework based on the generative diffusion model, named Joint Conditional Diffusion Model-based Inverse Problem Solver (JCDI).
In this framework, we train the conditional diffusion model to learn the inherent distributions within parameter space and provide a data-driven solution for parameter estimation. 
The key innovations of this work include:

\begin{enumerate}
\item Distinct from traditional conditioning mechanisms, where diffusion models rely on labeled conditions, we condition the diffusion model on system observations and generate system parameters that are constrained by the observations.
\item The probabilistic nature of diffusion models enables them to produce a distribution of parameters consistent with the observations, allowing for studying the parameter \textcolor {cyan}{dependencies} and quantifying uncertainties.
\item We develop a denoising neural network architecture based on a transformer encoder. It processes model parameters as input tokens and uses its attention mechanism to learn parameter \textcolor {cyan}{dependencies}, as well as the relationships between these parameters and their corresponding observations. 
\item To address the parameter non-uniqueness problem, we introduce the multi-event joint conditioning mechanism: conditioning the diffusion model on multiple observations under various disturbances. It produces a posterior parameter distribution that satisfies various observation constraints simultaneously. This probability under joint conditions effectively reduces parameter estimation uncertainties.
\end{enumerate}

With the study of CLM in power systems, we demonstrate a 58.6\% reduction in parameter estimation error with the increase of conditioned fault events.
Comprehensive comparison with existing RL- and SL-based methods and the modern inference baselines demonstrates that JCDI shows superior performance in addressing parameter non-identifiability, producing probabilistic solutions that achieve high accuracy in both parameter estimation and dynamic response prediction -- without requiring repeated simulations.  
\textcolor {cyan}{Beyond power system modeling, this approach can be extended to mitigate the parameter non-uniqueness challenge in diverse dynamical systems, leveraging its data-driven nature and probabilistic inference with the ability to handle multiple conditional observations.
This generality, coupled with substantial improvements over existing methods,} suggests that \thename could become a valuable tool for solving complex parameter identification problems across scientific disciplines.

\section*{Results}\label{sec2}

\bmhead{Simulation settings}

We adopt the advanced CLM -- the composite load model with distributed generation (CMPLDWG)~\citep{CMPLDW-DG} developed by Western Electricity Coordinating Council. This state-of-the-art model incorporates the increasingly penetrated power electronic-interfaced loads, single-phase induction motors, and distributed energy resources. The CMPLDWG model details are shown in Supplementary Section~1.
By applying electrical fault-induced disturbances in the transmission system, we generate the dynamic responses of CMPLDWG, including the transient trajectories of active and reactive powers measured at the point of interconnection.
Specifically, three distinct electrical faults with different fault clearing times and bus locations (Supplementary \textcolor {cyan}{Table~1}) are selected for the simulation study, including one that results in power electronic load tripping (trip fault), one which induces motor D stalling (stall fault), and another that exhibits neither phenomenon (ordinary fault).
 
CMPLDWG contains nearly 200 parameters.
Some demonstrate significant impact on the system dynamics, while others have minor effects and are difficult to estimate. To solve the parameterization problem efficiently, we conduct global sensitivity analysis using Sobol's method.
The analysis details and key findings are presented in Supplementary Section~3, which reveal notable sensitivity variations across different fault events.
Based on the parameter rankings according to the total Sobol indices under various fault events (Supplementary Figure~6), we select 30 parameters with relatively high sensitivity for identification. Supplementary Table~2 lists these parameters along with their notations and physical meanings.
We generate training and testing datasets by perturbing these sensitive parameters and running grid simulations under different fault-induced disturbances. \textcolor {cyan}{Supplementary Section~4.1} illustrates the data generation process.
 
\bmhead{Parameter estimation uncertainties}  
We train our diffusion-model-based parameterization framework to estimate system parameters from observed dynamic responses under grid fault-induced disturbances.
The algorithm implementation is explained in \textcolor {cyan}{Supplementary Section~4.1}. To evaluate the effectiveness of multi-event joint conditioning, we compare JCDI conditioned on three fault events, including ordinary, trip, and stall faults, with CDI (Conditional Diffusion Model-based Inverse Problem Solver) conditioned solely on the ordinary fault.

\begin{figure}[h]
	\centering
    \includegraphics[width=0.95\textwidth]{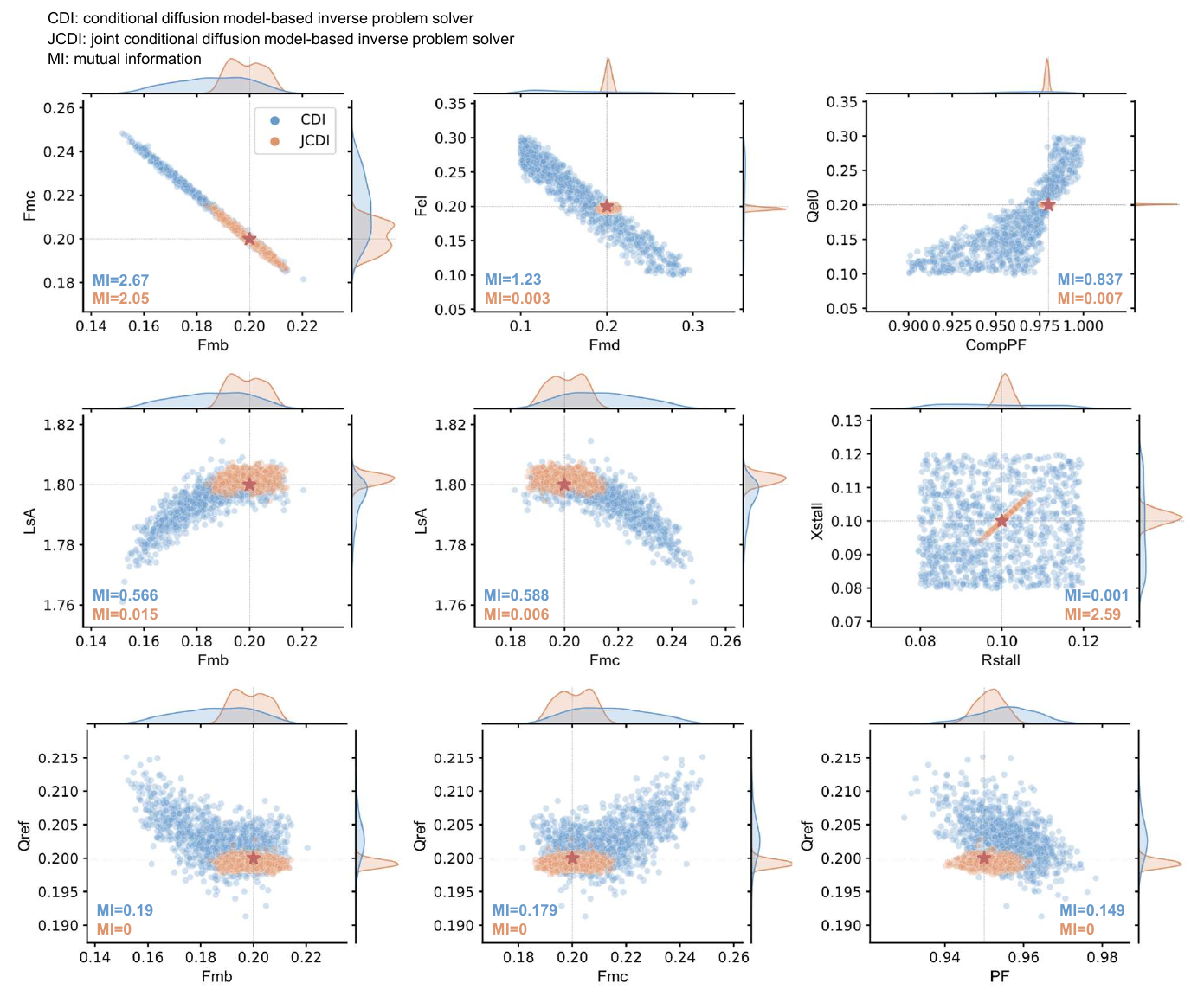} 
    \caption{\textbf{Representative cases of parameter posterior results.}
    The scatter plots reveal \textcolor {cyan}{associations} among parameter pairs.
    The marginal distribution for each model parameter is displayed along the top and right edges, demonstrating JCDI produces more concentrated posterior distributions for the non-identifiable parameters.
   Blue and orange dots represent parameters estimated by CDI and JCDI, respectively, with pink stars indicating the actual values.
   The statistical figures are constructed with 1000 data samples.}
	\label{fig:correlation-cmp}
\end{figure}

We inject the desired power trajectories into the well-trained diffusion models and deduce the posterior results of parameter estimation (Supplementary Figure~8). 
Some representative cases are displayed in Figure~\ref{fig:correlation-cmp}.
One thousand samples of parameter estimates are generated given the desired power trajectories, which are obtained from dynamic simulation with the default model parameters listed in Supplementary Table~2.

Parameter non-identifiability can arise due to several reasons. For example, some parameters may have vanishingly low sensitivity under specific conditions, making them practically unidentifiable from the available data. Meanwhile, other model parameters exhibit complex dependencies on each other. Then, different combinations of parameters that satisfy a specific mathematical relation can yield identical model outputs~\cite{Practical-identifiability}.
We use mutual information (MI) to quantitatively evaluate parameter dependencies~\cite{book-MI,6790247}. MI quantifies how much information one random variable provides about another. It can be calculated by Kullback–Leibler divergence between the joint distribution and the product of the marginal distributions of the two variables as expressed by equation~\eqref{MI}~\cite{PhysRevE-MI,scikit-MI}. Calculation results are presented in Figure~\ref{fig:correlation-cmp}.
\begin{equation}
I\left( {X;Y} \right) = \int_{\mathop{\rm y}\nolimits}  {\int_{\mathop{\rm x}\nolimits}  {p\left( {x,y} \right)\log \left( {\frac{{p\left( {x,y} \right)}}{{p\left( x \right)p\left( y \right)}}} \right)dxdy} }, 
\label{MI}
\end{equation}
where ${p\left( {x,y} \right)}$ is the joint probability density function of $X$ and $Y$ and ${p\left( x \right)}$ and ${p\left( y \right)}$ respectively represent the marginal probability density functions of $X$ and $Y$.
 
Our findings reveal notable \textcolor {cyan}{associations} among several sets of model parameters estimated by CDI. 
The load fractions ${F_{{\rm{mb}}}}$ and ${F_{{\rm{mc}}}}$, ${F_{{\rm{md}}}}$ and ${F_{{\rm{el}}}}$ are negatively correlated, yielding high MIs greater than 1 nat. The power factor of motor D (${CompPF}$) and the initial reactive power of electronic load (${Q_{{\rm{el0}}}}$) exhibit positive correlation, with an MI of 0.837 nat.  
In addition, we also observe notable \textcolor {cyan}{associations} between parameter pairs (${F_{{\rm{mb}}}}$ and ${L_{{\rm{sA}}}}$) and (${F_{{\rm{mc}}}}$ and ${L_{{\rm{sA}}}}$), showing MI values above 0.5 nat. 
These parameter combinations yield similar post-fault trajectories, validating their non-identifiability due to interdependency. 
Theoretically, these \textcolor {cyan}{dependencies} can be attributed to the counteracting contributions of steady states or dynamic behaviors associated with different load components, i.e., varied combinations of parameters may have opposite effects on the dynamic responses of individual loads, preserving the invariance of the total system response.
However, the MI between ${R_{{\rm{stall}}}}$ and ${X_{{\rm{stall}}}}$ is merely 0.001 nat, indicating no statistical \textcolor {cyan}{dependence}. CDI fails to identify ${R_{{\rm{stall}}}}$ and ${X_{{\rm{stall}}}}$ because they exhibit negligible sensitivity under the ordinary fault event (Supplementary Figure~6), where motor D stall does not occur.

When comparing JCDI with CDI, we find JCDI produces more concentrated posterior distributions with reduced variance for these non-identifiable model parameters.
For the \textcolor {cyan}{dependent} parameters, we notice their MIs are considerably reduced when estimated by JCDI. This implies JCDI effectively mitigates the parameter non-identifiability, as it introduces more constraints under different fault conditions.
As a result, ${F_{{\rm{md}}}}$, ${F_{{\rm{el}}}}$, ${Q_{{\rm{el0}}}}$, and $CompPF$ estimated by JCDI closely approximate the true values.
For the insensitive parameters, they exhibit markedly enhanced sensitivity under the additional faults with the occurrence of motor D stalling and electronic load tripping. The enhanced sensitivity facilitates their identification.
Therefore, JCDI substantially narrows the posterior distributions of model parameters that fail to be estimated by CDI, such as ${R_{{\rm{stall}}}}$, ${X_{{\rm{stall}}}}$, ${f_{{\rm{rcel}}}}$, ${N_{{\rm{q1}}}}$, and ${N_{{\rm{q2}}}}$ \textcolor {cyan}{(Supplementary Figure~8)}.
Notably, ${R_{{\rm{stall}}}}$ and ${X_{{\rm{stall}}}}$ show a strong positive correlation with a high MI of 2.59 nat when estimated by JCDI. This correlation appears with the occurrence of motor D stalling. More constraints would be required for mitigation.

When evaluating parameter estimation accuracy, mean absolute percentage error is commonly used to measure relative precision~\citep{de-Myttenaere-2016}. However, it becomes invalid when the actual parameter value equals zero.  
To address this limitation, we define range percentage error (RPE) for an individual model parameter as formulated in equation~\eqref{RPE}. We then use mean absolute range percentage error (MARPE), the average of RPEs across different model parameters, to assess the overall estimation accuracy. 
In both RPE and MARPE calculations, the absolute parameter errors are normalized by the parameter ranges rather than the actual parameter values, ensuring valid and consistent measurements across all parameter values regardless of their positions within the specified range.

\begin{equation}
RP{E_i} = 100\left| {\frac{{{{\hat \theta }_i} - {\theta _{i,0}}}}{{U{B_i} - L{B_i}}}} \right|  \label{RPE},
\end{equation}

\noindent where ${\hat \theta }$ and ${{\theta _0}}$ respectively represent the estimated and actual values of parameters, $UB$ and $LB$ denote the upper and lower bounds, and $i$ is the parameter index. 

\begin{figure}[!htp]
	\centering
    \includegraphics[width=0.9\textwidth]{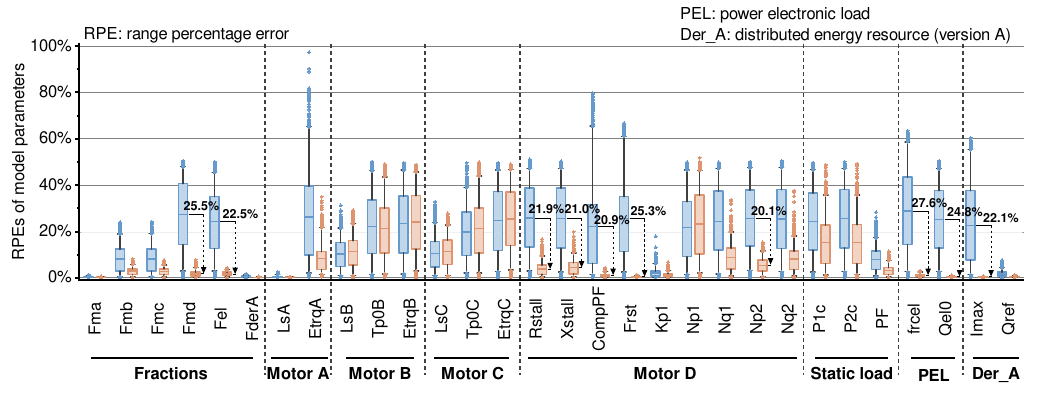} 
    \caption{\textbf{Range percentage errors of estimated model parameters.} 
    RPEs are calculated using equation~\eqref{RPE} for each model parameter across 1000 data samples and visualized through the box plot.
    \textcolor {cyan}{The calculation results for CDI and JCDI are shown in blue and orange, respectively.}
    The boxes cover the intervals from the 25th to 75th percentiles 
    with mean values shown as horizontal lines.
    The lower and upper ends of the whiskers respectively represent the 5th and 95th percentiles, while
 the dots beyond these bounds indicate the outliers. }
\label{fig:parameter-error}
\end{figure}

Figure~\ref{fig:parameter-error} presents the calculation result of RPE.
Some of the model parameters, e.g., ${F_{{\rm{md}}}}$, ${F_{{\rm{el}}}}$, ${f_{{\rm{rcel}}}}$, and ${Q_{{\rm{el0}}}}$, exhibit high uncertainties when estimated by CDI. 
The mean MARPE across all parameter estimate samples is 18.0\%.
Nevertheless, JCDI significantly reduces the estimation errors for multiple model parameters, particularly for those that present higher sensitivity under the trip and stall faults, such as ${F_{{\rm{el}}}}$, ${F_{{\rm{md}}}}$, ${f_{{\rm{rcel}}}}$, ${R_{{\rm{stall}}}}$, ${X_{{\rm{stall}}}}$, and ${F_{{\rm{rst}}}}$, as well as the \textcolor {cyan}{dependent} parameters ${F_{{\rm{mb}}}}$, ${F_{{\rm{mc}}}}$, ${Q_{{\rm{el0}}}}$, and $CompPF$.
As a result, the \textcolor {cyan}{mean} MARPE decreases to 7.46\%.
Together, these findings suggest that with multi-event joint conditioning, JCDI effectively reduces the parameter uncertainties, yielding more accurate parameter estimates.

To further investigate CMPLDWG's modal property, we compare the eigenvalue distributions for the system with actual parameters, and with parameters estimated by JCDI (Supplementary Figure~9). 
We observe CMPLDWG's eigenvalues with parameter estimates deduced by JCDI locate closely to the actual eigenvalues.
We evaluate the eigenvalues' accuracy using the relative error of their moduli.
The results indicate that relative errors of eigenvalue moduli remain within 5.23\% under different parameter estimates with a mean relative error of 0.35\%.
This validates the consistency in modal properties between the system with estimated parameters and the actual system.

\begin{figure}[h]
	\centering
    \includegraphics[width=0.95\textwidth]{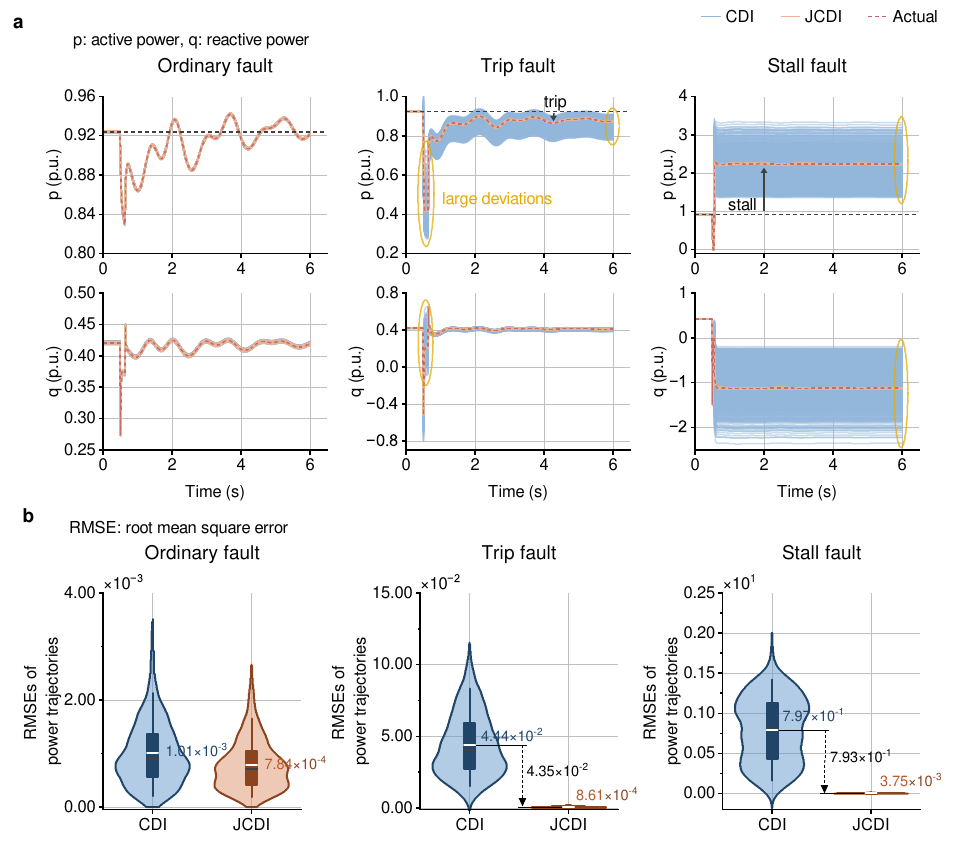} 
	\caption{\textbf{Prediction results of dynamic responses.} 
    \textbf{a} Predicted power trajectories, including active power (p) and reactive power (q), under different fault events.
    They are obtained via grid simulations with estimated model parameters.
    \textcolor {cyan}{The blue and orange lines represent the predicted power trajectories of CDI and JCDI, respectively. The pink dashed lines are the actual power measurements. The yellow annotations highlight prediction deviations.}
    \textbf{b} RMSEs of predicted power trajectories. The violin plots show the probability distributions of RMSEs with the widths representing the approximate frequency of data points in each region. The boxes inside span the intervals from the 25th to 75th percentile, while the white and black lines correspondingly represent the mean and median values. \textcolor {cyan}{The calculation results for CDI and JCDI are shown in blue and orange, respectively.
    The plots are constructed using 1000 data samples.}
    }
	\label{fig:traj-result}
\end{figure}

\bmhead{Prediction of dynamic responses} 
Next, we analyze the prediction results of dynamic responses with the estimated model parameters. 
Substituting the estimated model parameters into CMPLDWG, we perform transient simulations under different fault-induced disturbances with the resulting power trajectories shown in Figure~\ref{fig:traj-result}a. 
To assess prediction accuracy, we calculate the root mean \textcolor {cyan}{square} error (RMSE) between the simulated and actual power trajectories as formulated in equation~\eqref{RMSE}. Figure~\ref{fig:traj-result}b shows the RMSE results.

\begin{equation}
RMSE = \sqrt {\sum\nolimits_{t = 1}^T {\frac{{{{\left( {{p_{\hat \theta }}\left( t \right) - {p_0}\left( t \right)} \right)}^2}}}{T} + } } \sqrt {\sum\nolimits_{t = 1}^T {\frac{{{{\left( {{q_{\hat \theta }}\left( t \right) - {q_0}\left( t \right)} \right)}^2}}}{T}} } \label{RMSE},
\end{equation}
\noindent where ${{p_{\hat \theta }}\left( {t} \right)}$ and ${{q_{\hat \theta }}\left( {t} \right)}$ respectively represent the estimated active and reactive powers at time $t$ , ${{p_0}\left( {t} \right)}$ and ${{q_0}\left( {t} \right)}$ are the actual active and reactive powers, and $t$ is the time instant.
 
As a form of data regularization, we initiate the electrical faults at a common time of 0.5 s from the system's steady states. Therefore, the power trajectories well encompass the pre-fault steady-state conditions, the transient responses during the faults, and the post-fault dynamic recovery processes.
When the electric faults occur, both active and reactive powers dip sharply. For the ordinary fault event, the powers oscillate dynamically and recover to their initial values after fault clearance.
However, the steady-state powers settle below the initial value due to incomplete load recovery with electronic load tripping, and the absolute values of steady powers increase markedly after the fault with motor D stalling. 
Despite parameter estimation uncertainties, both CDI and JCDI accurately predict power trajectories for the ordinary fault, achieving mean RMSEs of approximately $1 \times {10^{ - 3}}$.
Nevertheless, we observe substantial deviation between the power trajectories predicted by CDI and the actual ones for the trip and stall faults with the mean RMSEs increasing to $4.44 \times {10^{ - 2}}$ and $7.97 \times {10^{ - 1}}$, respectively.
In contrast, JCDI maintains high prediction accuracy for these two faults with respective mean RMSEs of $8.61 \times {10^{ - 4}}$ and $3.75 \times {10^{ - 3}}$. 
These results reveal the limitations of parameter estimation using a single disturbance, where some solutions fail to replicate system dynamic behavior under additional disturbances.
JCDI achieves consistent performance across different fault-induced disturbances by successfully constraining the non-identifiable parameters.

\begin{figure}
	\centering
    \includegraphics[width=0.9\textwidth]{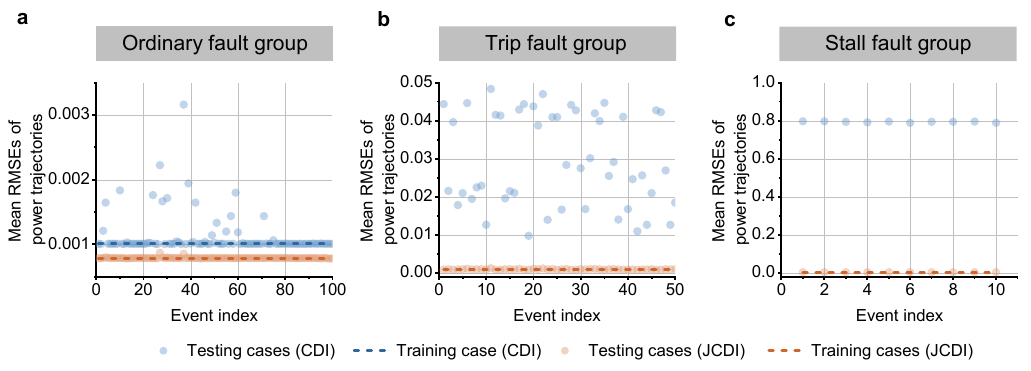} 
\caption{\textbf{Generalization results for additional testing fault events.} Mean RMSEs of power trajectories are calculated using 1000 parameter estimate samples for each testing fault event. Blue and orange dots represent the mean RMSEs produced by CDI and JCDI respectively across different testing events. Dashed lines indicate the mean RMSEs for training events: \textbf{a} the ordinary fault event group, \textbf{b} the load tripping fault event group, and \textbf{c} the motor stalling fault event group.}
	\label{fig:test-RMSEs}
\end{figure}

\bmhead{Out-of-distribution generalization evaluation}  
While the model parameters are derived from observed dynamic responses under specific fault-induced disturbances during training, they are also expected to predict responses under unencountered disturbances accurately.
This refers to out-of-distribution generalization, i.e., the ability to generalize to datasets with different distributions from the training data~\citep{liu2023outofdistributiongeneralizationsurvey}.
We test the out-of-distribution generalization performance of parameter estimates using additional fault scenarios with randomized bus locations and fault clearing times. These scenarios comprise three groups: 100 ordinary fault events, 50 fault events inducing power electronic load tripping, and 10 events with motor D stalling.
Figure~\ref{fig:test-RMSEs} presents the test results. 
For CDI, the mean RMSEs of power trajectories for most testing cases within the ordinary fault group are close to the training case, while increased errors appear in several events, such as event 37.  
However, its performance deteriorates, exhibiting dispersed errors for tripping fault events and substantial increases in RMSE for stalling fault events.  
In contrast, JCDI maintains consistent and accurate predictions across all three testing groups.
These results further demonstrate that JCDI produces parameter estimates with favorable generalization capability.

\bmhead{\textcolor {cyan}{Impact of Data Quality}} 
The real measurements from phasor measurement units contain noise and incomplete data~\cite{6111219}. Model robustness should be guaranteed for successful real-world application.
Here, we evaluate the influence of noise and dropout on our model performance.
We construct noisy synthetic data by adding white Gaussian noise to the simulated measurements as formulated by equation~\eqref{noisy-data}.
\begin{equation}
\begin{array}{l}
{p_{\theta {\rm{,noisy}}}}\left( t \right){\rm{ = }}{p_\theta }\left( t \right) + \varepsilon \left( t \right)\\
{q_{\theta {\rm{,noisy}}}}\left( t \right){\rm{ = }}{q_\theta }\left( t \right) + \varepsilon \left( t \right)
\end{array} \label{noisy-data},
\end{equation}
\noindent where $\varepsilon \left( t \right)$ is the Gaussian noise at time instant $t$ and ${p_{\theta ,{\rm{noisy}}}}\left( t \right)$ and ${q_{\theta ,{\rm{noisy}}}}\left( t \right)$ respectively represent the noisy active and reactive powers.

{
\color{cyan}
Considering a wide range of noise levels~\cite{7741972,10620390}, we corrupt the training data using Gaussian noise with a signal-to-noise ratio (SNR) between 45 and 75 dB, retrain the model, and evaluate its performance at different noise levels.
Using a 50 dB SNR as an example, the parameter estimation results are shown in Supplementary Section~4.2.
The existence of measurement noise aggravates the ill-conditioning of the parameterization problem, making it more challenging for accurate parameter identification~\citep{dissertation-noise}. 
The correlations between ${F_{{\rm{mb}}}}$ and ${F_{{\rm{mc}}}}$, ${F_{{\rm{md}}}}$ and ${F_{{\rm{el}}}}$ estimated by CDI become inapparent (Fig.~\ref{fig:posterior-result-noisy}).
JCDI still effectively reduces the parameter estimation uncertainties and produces concentrated posterior distributions for model parameters, such as ${F_{{\rm{md}}}}$, ${F_{{\rm{el}}}}$, ${CompPF}$, and ${Q_{{\rm{el0}}}}$.
Despite an increase of parameter estimation uncertainties, both CDI and JCDI achieve comparable performance in dynamic response prediction with the previous noise-free condition. 
The credible intervals of power trajectories predicted by JCDI are close to the envelopes of the noisy measurements under different fault-induced disturbances, highlighting its high fidelity in preserving the load dynamics (Supplementary Figure~11).
We further investigate the sensitivity of our model to noise level.
Within a 45–75 dB SNR range, JCDI's mean parameter MARPE remains below 16\% (Supplementary Figure~12).
Results demonstrate the robustness of our model across different noise levels despite moderate performance degradation. 
The parameter non-identifiability challenge becomes increasingly severe in the presence of measurement noise, which necessitates effective data filtering techniques, as well as advanced mitigation strategies.

\begin{figure}
	\centering
    \includegraphics[width=0.95\textwidth]{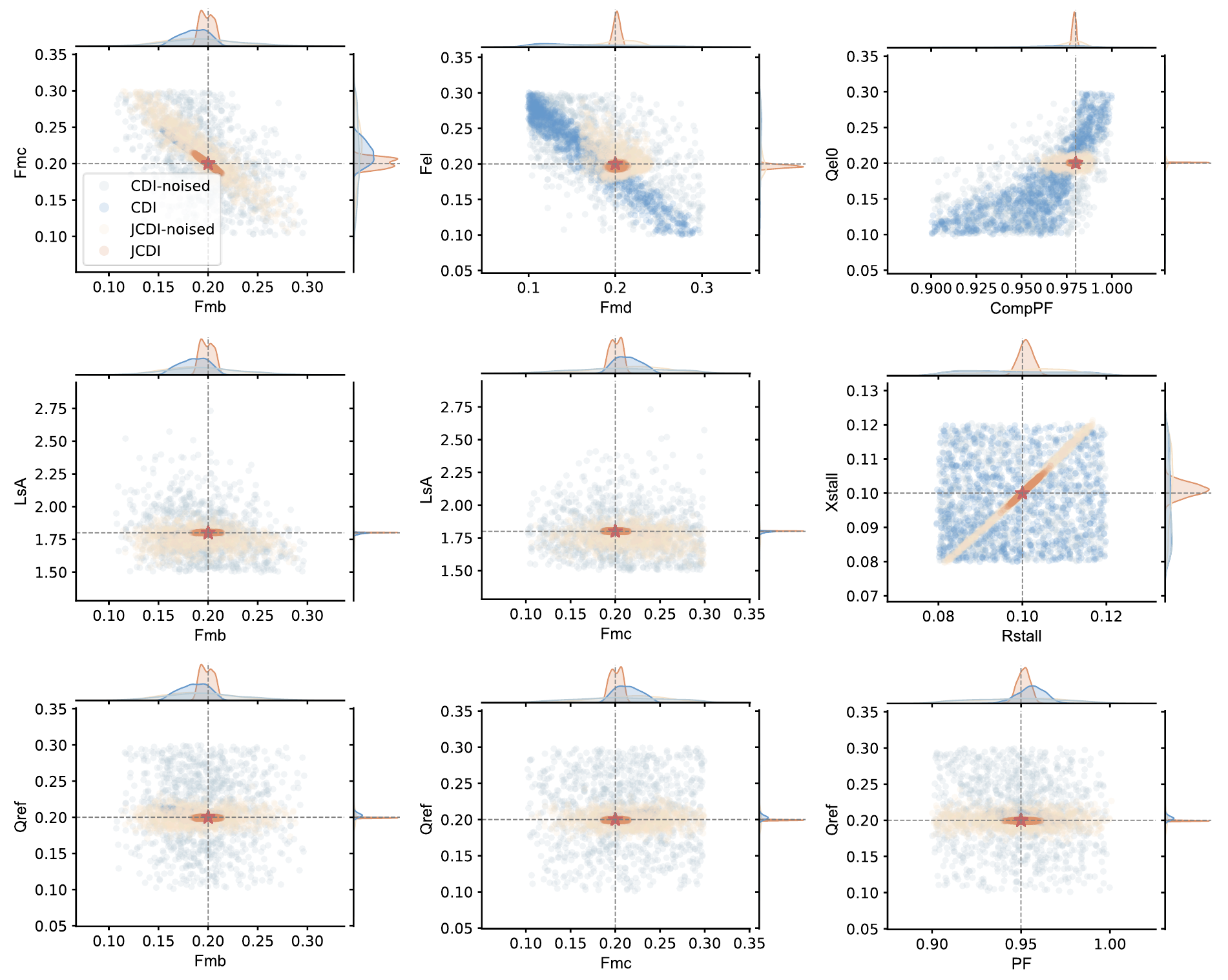} 
\caption{\textbf{Representative cases of parameter posterior results considering measurement noise at 50 dB SNR.}    
    The scatter plots of parameter pairs are displayed inside, with the marginal distribution for each model parameter presented along the top and right edges.
   Blue and orange dots respectively represent parameters estimated by CDI and JCDI with noise-free measurements, while green and yellow dots respectively denote parameter estimates of CDI and JCDI with noisy measurements.
   Pink stars indicate the actual values.
   The statistical figures are constructed with 1000 data samples.}
	\label{fig:posterior-result-noisy}
\end{figure}

To evaluate model performance with incomplete data, we reconstruct synthetic data with dropout by randomly removing 10\% of the data points in the active and reactive power trajectories.
Detailed results are provided in Supplementary Section~4.2.
Similar to the findings on noise impact, JCDI-based parameters still accurately replicate the dynamic responses when inferring with the incomplete data while providing parameter estimates with modestly higher uncertainty. 
The mean parameter MARPE becomes 13.6\%, while the mean trajectory RMSEs under the ordinary and trip faults measure around $2 \times {10^{ - 3}}$ and remain within $1 \times {10^{ - 2}}$ under the stall fault. 
}

{
\color{cyan}
\bmhead{Comparison with existing parameterization methods} 

We compare JCDI with two mainstream methods for CLM parameterization and modern parameter inference baselines, including:
(1) RL implemented by Deep Q-Network (DQN); 
(2) SL using a combined neural network architecture, including a Residual Network (ResNet) in series with a transformer encoder, designated as \textit{Res-TFR}.
This architecture matches JCDI's neural network structure but infers model parameters directly, allowing the comparison to serve as an ablation study that demonstrates the advantage of the diffusion process;
(3) Flow-Matching Posterior Estimation (FMPE)~\citep{wildberger2023flow};
(4) Approximate Bayesian Computation-Sequential Monte Carlo (ABC-SMC)~\citep{ABC-SMC}; and 
(5) Covariance Matrix Adaptation Evolution Strategy (CMA-ES)~\citep{hansen2006eda}.

Among these parameterization methods, Res-TFR learns the mapping between observations and model parameters, producing deterministic results.  
CMA-ES represents an advanced evolutionary algorithm for numerical optimization of nonlinear problems, where the best solution within the population is used for evaluation. 
DQN, ABC-SMC, FMPE, and JCDI all yield probabilistic solutions. 
Each algorithm is trained under both a single-fault event condition (SEC) and multi-fault event condition (MEC). 
Algorithm implementations are described in Supplementary Section~5.1 with the corresponding results in Supplementary Section~5.2.
We perform a comprehensive comparison of different parameterization methods in terms of identifiability, calibration, and inference efficiency, summarized in Table~\ref{tab:cmp-algs-terms}. 

\begin{table}[h]
	\caption{Comprehensive comparison of different parameterization methods}
	\centering
	\begin{tabular}{lllll}
		\toprule
		\multirow{2}{*}{\makecell[l]{Parameterization\\ methods}} & \multirow{2}{*}{\makecell[l]{Calibration}} &\multirow{2}{*}{\makecell[l]{Identifiability}}  & \multirow{2}{*}{\makecell[l]{Inference efficiency}}\\\\
		\midrule
		DQN         &  $\star \star $ &  $\star \star $    &  $\star $ \\ \hline
		Res-TFR      &  $\star  \star  \star  \star $  &  $\star  \star  \star  \star$   &   $\star  \star \star  $  \\ \hline
		CMA-ES       &  $\star  \star  \star \star  $  &  $\star \star \star $   &  $\star $  \\ \hline        
		ABC-SMC      &  $\star \star $  &  $\star \star $   &  $\star \star $  \\ \hline
		FMPE         &  $\star  \star  \star $  &  $\star  \star  \star $      &  $\star  \star  \star  \star \star $ \\ \hline
		JCDI         & $\star  \star  \star  \star \star $     &  $\star  \star  \star  \star \star$    &  $\star  \star  \star \star  $    \\ 
		\bottomrule
	\end{tabular}
	\label{tab:cmp-algs-terms}
\end{table}

For the identifiability issue, JCDI effectively reduces estimation uncertainties for non-identifiable parameters under MEC, while Res-TFR provides accurate parameter estimates for most of the parameters, including those exhibiting interdependencies. 
With the assistance of MEC, parameter estimation uncertainties are also reduced for FMPE.
The posterior distributions of ${F_{{\rm{md}}}}$, ${F_{{\rm{el}}}}$, $CompPF$, and ${Q_{{\rm{el0}}}}$ become much more concentrated than those estimated using a single-fault event (Supplementary Fig.~17). However, FMPE still fails to accurately identify parameters such as ${F_{{\rm{mb}}}}$, ${F_{{\rm{mc}}}}$, and ${L_{{\rm{sA}}}}$.
Minor improvements in parameter identifiability are observed for DQN and ABC-SMC with multi-event learning. 
For ABC-SMC, MEC helps reduce the estimation uncertainties of parameters ${F_{{\rm{md}}}}$, ${R_{{\rm{stall}}}}$, and ${X_{{\rm{stall}}}}$, which present high sensitivity under the stall fault, but it fails to work for other parameters.
Large parameter deviations are observed for DQN under both SEC and MEC, while ${F_{{\rm{md}}}}$ gets closer to its actual value when DQN is trained with multiple faults.

In terms of calibration, we present the parameter inference results under MEC in Supplementary Figure~21, and show the results of dynamic response prediction in Supplementary Figure~22. We then calculate the evaluation metrics of estimation accuracy in Table~\ref{tab:cmp-accuracy-algs}, including parameter MARPEs and trajectory RMSEs under different fault disturbances. Solutions that minimize trajectory RMSEs under the ordinary fault in the solution sets are denoted as the min-traj cases.
DQN and ABC-SMC exhibit low estimation accuracy.
DQN tends to converge to local optima. Both the parameter estimates and predicted power trajectories show evident deviations.
When trained under SEC, it achieves the minimum trajectory RMSE at $5.30 \times {10^{ - 3}}$ for the ordinary fault with the MARPE of model parameters exceeding 20\%. 
Unexpectedly, the estimation errors become larger under MEC.
Different fault events are selected randomly for each episode during multi-event training, which may lead to a non-stationary environment and degrade the algorithm performance.
ABC-SMC produces parameter estimates with high uncertainties and presents wide credible intervals of predicted power trajectories.
When trained under MEC, ABC-SMC effectively reduces the prediction uncertainty under the stall fault at the expense of increasing prediction errors under the ordinary and trip faults (Supplementary Figure~22b).
Res-TFR achieves the lowest parameter MARPE of 5.23\% under MEC while maintaining relatively high RMSEs of power trajectories, which aligns with its loss function. 
In contrast, CMA-ES exhibits excellent performance in the prediction of power trajectories, but it fails to accurately estimate the model parameters.
Under multi-event training, the trajectory RMSEs of CMA-ES under different fault events are below $2 \times {10^{ - 3}}$ with the parameter MARPE decreasing slightly to 16.5\%.
Obvious deviations are observed for parameters estimated by CMA-ES (Supplementary Figure~21), such as ${F_{{\rm{md}}}}$, ${F_{{\rm{el}}}}$, ${L_{{\rm{sA}}}}$, ${R_{{\rm{stall}}}}$, and ${X_{{\rm{stall}}}}$, some of which are dependent as previously demonstrated. 
CDI and JCDI probabilistically generate solution sets that encompass various combinations of model parameters through the diffusion process. This increases the parameter MARPE compared with Res-TFR. 
The posterior distributions for JCDI-based parameters, such as ${F_{{\rm{mb}}}}$ and ${E_{{\rm{trqB}}}}$, contain multiple peaks (Supplementary Figure~21), which can be caused by the parameter dependencies and the probabilistic nature of JCDI.
With the assistance of joint conditioning, JCDI achieves a mean parameter MARPE of 7.46\%, representing a 58.6\% reduction compared to CDI, while delivering high trajectory estimation accuracy across various fault events with respective mean RMSEs below $4 \times {10^{ - 3}}$. 

\begin{table}[h]
	\caption{\textcolor {cyan}{Evaluation of estimation accuracy for different parameterization methods}}
	\centering
	\begin{tabular}{lllll}
		\toprule
		\multirow{2}{*}{\makecell[l]{Parameteri-\\ zation methods}} & \multirow{2}{*}{\makecell[l]{MARPEs of\\ parameters}} &\multicolumn{3}{c}{RMSEs of power trajectories}   \\
		\cmidrule(r){3-5}
		& & Ordinary fault & Trip fault  &  Stall fault \\
		\midrule
		DQN-SEC (mean)       &  23.4\%  & $6.46 \times {10^{ - 3}}$   & $2.44 \times {10^{ - 2}}$   & $1.21 \times {10^{ - 1}}$   \\
		DQN-SEC (min-traj)   &  23.5\%  &  $5.30 \times {10^{ - 3}}$   &  $2.38 \times {10^{ - 2}}$    &   $1.55 \times {10^{ - 1}}$   \\ \hline
		DQN-MEC (mean)       &  24.8\%  & $2.20 \times {10^{ - 2}}$   & $4.73 \times {10^{ - 2}}$   & $1.64 \times {10^{ - 1}}$   \\ \Xhline{0.6pt}
		Res-TFR-SEC           &  10.6\%  &  $3.55 \times {10^{ - 3}}$   &  $2.21 \times {10^{ - 2}}$    &     $6.43 \times {10^{ - 1}}$    \\ \hline
		Res-TFR-MEC           &  5.23\%  &  $5.46 \times {10^{ - 3}}$   &  $5.35 \times {10^{ - 3}}$    &     $7.46 \times {10^{ - 3}}$    \\     
		\Xhline{0.6pt}
		CMA-ES-SEC           &  22.1\%  &  $1.21 \times {10^{ - 4}}$   &  $5.74 \times {10^{ - 2}}$    &     $6.98 \times {10^{ - 1}}$    \\  \hline
		CMA-ES-MEC           &  16.5\%  &  $7.44 \times {10^{ - 4}}$    &  $1.73 \times {10^{ - 3}}$    &     $1.76 \times {10^{ - 3}}$   \\  \Xhline{0.6pt}
		ABC-SMC-SEC (mean)           &  23.4\%  &  $5.80 \times {10^{ - 3}}$   &  $3.80 \times {10^{ - 2}}$    &     $6.93 \times {10^{ - 1}}$    \\  \hline
		ABC-SMC-MEC (mean)           &  23.2\%  &  $3.26 \times {10^{ - 2}}$    &  $4.39 \times {10^{ - 2}}$     &     $6.17 \times {10^{ - 2}}$     \\  \Xhline{0.6pt}
		FMPE-SEC (mean)       &  21.1\%  &  $2.33 \times {10^{ - 3}}$   &  $4.37 \times {10^{ - 2}}$    &     $7.18 \times {10^{ - 1}}$    \\  \hline
		FMPE-MEC (mean)       &  16.1\%  &  $2.55 \times {10^{ - 3}}$   &  $3.43 \times {10^{ - 3}}$    &     $1.53 \times {10^{ - 2}}$    \\  \Xhline{0.6pt}
		CDI (mean)        &  18.0\%  & $1.01 \times {10^{ - 3}}$    & $4.44 \times {10^{ - 2}}$    &    $7.97 \times {10^{ - 1}}$   \\
		CDI (min-traj)    &  19.8\%  &   $3.24 \times {10^{ - 5}}$    &  $7.81 \times {10^{ - 2}}$      &  1.06          \\ \hline
		JCDI (mean)         & 7.46\%   &  \textbf{$7.84 \times {10^{ - 4}}$}    &  \textbf{$8.61 \times {10^{ - 4}}$}    &    $3.75 \times {10^{ - 3}}$   \\
		JCDI (min-traj)     & 8.99\%   & $4.18 \times {10^{ - 5}}$    &  $3.44 \times {10^{ - 4}}$   &      $5.47 \times {10^{ - 3}}$  \\
		\bottomrule
	\end{tabular}
	\label{tab:cmp-accuracy-algs}
\end{table}

With regard to calculation efficiency, DQN, CMA-ES, and ABC-SMC require performing transient simulations of the power system model repeatedly during both training and inference processes, which is time-consuming.
In addition, the training process merely allows the algorithm to learn the parameter estimates that approximate one specific observation. 
However, for Res-TFR, FMPE, and JCDI, we conduct parallel simulations to generate the training and testing datasets. Once completed, there is no need for additional physical simulation in both training and inference processes. 
Furthermore, they all learn the mapping between different observations and model parameters, providing a general inverse solution of the system dynamics.
It is also worth noting that ABC-SMC, FMPE, and JCDI probabilistically deduce multiple parameter estimates during the inference stage.
Nevertheless, Res-TFR provides only one solution.

}

\section*{Discussion}\label{sec13}

This work presents JCDI, a probabilistic parameter estimation framework that mitigates 
the critical challenges posed by parameter non-uniqueness and cross-event generalization. Successful verification of JCDI has been achieved for CLM in power systems. The comprehensive evaluation yields several key findings, including: 

\begin{itemize}[leftmargin=*]
 \item The global sensitivity analysis reveals how the model parameters influence the system's dynamic behavior differently under various grid faults. This insight inspires our approach to identify model parameters using multiple fault-induced disturbances.
\item When conditioned on a single disturbance, the diffusion model generates a probabilistic solution encompassing multiple parameter combinations. They accurately reproduce power trajectories for the specific disturbance yet fail to generalize across diverse fault-induced disturbances, demonstrating the inherent non-identifiability of CLM parameters.
\item  JCDI produces more concentrated posterior distributions for model parameters, effectively addressing non-identifiable cases and improving parameter estimation accuracy. Moreover, the model parameters estimated by JCDI accurately reproduce power trajectories under a group of testing fault events, verifying the favorable out-of-distribution generalization performance.
\item  Comparative studies demonstrate the superiority of diffusion models in solving the parameterization inverse problem compared to existing approaches. Diffusion models independently generate probabilistic solutions that accurately predict power trajectories and enable more precise and generalizable parameter estimation through the joint conditioning mechanism. 
In addition, they eliminate the need for iterative forward simulations, providing more computationally efficient solutions.
\end{itemize}

\subsection*{Limitations and future work}
Building on the effective demonstration of JCDI for load modeling, potential extensions may involve parameterization for other electrical systems, such as power electronic converters and energy storage systems, along with broader applications beyond electrical engineering and real-world deployment. 
However, several limitations warrant further discussion and represent important directions for future research.

\textbf{\textcolor {cyan}{Scalability considerations.}}
Empirical evidence from diffusion models~\cite{DDPM} and transformer architectures~\cite{touvron2023llamaopenefficientfoundation} suggests the JCDI framework should scale effectively to systems with hundreds to low thousands of parameters while ensuring computational tractability and training stability.
Therefore, it potentially can be applied to more complex systems, e.g., multi-area power grids.
In addition, it would be beneficial to reduce the computational requirements for large-scale implementations by exploring mitigation strategies, such as subsystem division and using task-specific compact latent spaces~\cite{hinton2006reducing,alemi2017deep}.

\textbf{Time-critical applications.}
The average inference time for JCDI on a NVIDIA GeForce RTX 3090 GPU is measured at 1107 ms per batch (128 samples). 
Nevertheless, several well-established acceleration strategies can be readily applicable to our architecture.
These include diffusion model subsampling techniques (e.g., \textcolor {cyan}{denoising diffusion implicit models}~\cite{song2021denoising}), lower-precision arithmetic~\cite{micikevicius2018mixed} and model distillation methods~\cite{salimans2022progressive}.
Together, these optimizations could yield at least an order of magnitude improvement in inference speed, making the framework practical for time-critical applications.

\textbf{Discontinuous nonlinearity.}  
CLM considered in our study represents an average model, despite its high nonlinearity. 
When extending JCDI to other systems, for example power electronic systems, their massive switching behavior induces strong discontinuous nonlinearities, which must be thoroughly considered when modeling electromagnetic transient processes~\cite{power-electronic}.
In this context, incorporating the discrete event information as physical priors into the framework could be a promising approach to improve its robustness in non-smooth dynamics~\citep{hao2023physicsinformedmachinelearningsurvey,karniadakis2021physics,11358392}.

\textbf{\textcolor {cyan}{Real-world deployment.}} 
The existence of measurement noise and dropout aggravates the ill-conditioning of the parameterization problem, making it more difficult for accurate parameter identification.
Effective data pre-processing methods, such as data filtering and missing data imputation, should be investigated to improve the data quality.
Meanwhile, there will be discrepancies between simulation and real measurements due to the modeling bias of power loads and noise statistics, which may create a domain shift challenge~\cite{quinonero2022dataset}. 
Establishing more accurate noise models and developing domain shift mitigation strategies~\cite{Csurka2017, DBLP:journals/pieee/ZhuangQDXZZXH21} will be explored as future work for real-world deployment of our method.


\section*{Methods}
\label{sec:method}
{
\color{cyan}
\subsection*{Problem formulation}
}

The measurement-based parameterization for CLM represents a classical inverse problem, where model parameters are derived from observed dynamic responses as formulated in equation~\eqref{inverse}. 
\begin{equation}
\mathbf{y} = F\left( \bm{\mathrm{\theta}}  \right) \Rightarrow \bm{\mathrm{\theta}}  = {F^{ - 1}}\left( \mathbf{y} \right) \label{inverse},
\end{equation}
where $F$ is the forward operator, ${F^{ - 1}}$ is inverse operator, $\bm{\mathrm{\theta}}$ denotes the model parameters, and $\mathbf{y}$ represents the observation. 
The observation comprises active and reactive powers measured at the interconnection point between the transmission system and power loads (Fig.~\ref{fig:WECC-CLM}).
The observation vector $\mathbf{y}$ can be expressed as equation~\eqref{measurement}.
\begin{equation}
\begin{array}{l}
\mathbf{y} = \mathbf{[p,q]}{\rm{ }}\\
\mathbf{p} = [{p_1},{p_2}, \cdots ,{p_T}]{\rm{ }}\\
\mathbf{q} = [{q_1},{q_2}, \cdots ,{q_T}],
\end{array}\label{measurement}
\end{equation}
where $\mathbf{p}$ and $\mathbf{q}$ represent the active and reactive power trajectories, respectively.

Bayesian inference provides a probabilistic approach for solving this inverse problem.  
Based on Bayes' theorem~\citep{murphy2012machine}, the posterior distribution of model parameters, conditioned on the specific observation, is proportional to the product of the likelihood function and prior parameter distribution, which is formalized in equation~\eqref{Baysian}.
\begin{equation}
P\left( {\bm{\mathrm{\theta}} |{{\mathbf{y}_0}}} \right) = \frac{{P\left( {{{\mathbf{y}_0}}|\bm{\mathrm{\theta}} } \right) \cdot P\left( \bm{\mathrm{\theta}}  \right)}}{{P\left( {{\mathbf{y}_0}} \right)}} \propto P\left( {{{\mathbf{y}_0}}|\bm{\mathrm{\theta}} } \right) \cdot P\left( \bm{\mathrm{\theta}}  \right) \label{Baysian},
\end{equation}
where ${\mathbf{y}_0}$ denotes the conditioned observation and ${P\left( \bm{\mathrm{\theta}}  \right)}$ and $P\left( {\bm{\mathrm{\theta}} |{{\mathbf{y}_0}}} \right)$ respectively represent the prior and posterior distributions of model parameters. The likelihood function $P({\mathbf{y}_0}|\bm{\mathrm{\theta}})$ quantifies the probability of observing ${\mathbf{y}_0}$ given specific parameters $\bm{\mathrm{\theta}}$. ${P\left( {{\mathbf{y}_0}} \right)}$ represents the probability of observing ${\mathbf{y}_0}$ across all the model parameters, serving as a constant.

{
\color{cyan}
\subsection*{Model development}
}  

The denoising diffusion probabilistic model (DDPM) is an advanced latent variable model trained using variational inference~\citep{DDPM}. 
The posterior is approximated with a parametric family of distributions, eliminating the need to compute the likelihood function~\citep{nemani2023uncertainty}.
Inspired by non-equilibrium thermodynamics, DDPM transforms the original data ${\mathbf{x}_0} \sim q\left( {{\mathbf{x}_0}} \right)$ into a latent representation ${\mathbf{x}_T} \sim N\left( {\bm{\mathrm{\mu}} ,{\bm{\mathrm{\sigma}} ^2}} \right)$ by gradually adding random noise then learns a Markov chain ${p_\omega }\left( {{\mathbf{x}_{0:T}}} \right)$ to reverse this process.
The reverse process is implemented via a denoising neural network that estimates and progressively removes the added noise, enabling the reconstruction of the original data distribution through the transition from ${\mathbf{x}_T}$ back to ${\mathbf{x}_0}$.

As an extension, the conditional DDPM enables controlled data generation that satisfies specific constraints.
The conditioning is achieved by modifying the reverse process to depend on both the noisy input and the condition.
Specifically, the denoising neural network ${\bm{\mathrm{\varepsilon}}_\omega }({\mathbf{x}_t},t)$ is augmented to accept an additional condition input $\mathbf{y}$, becoming ${\bm{\mathrm{\varepsilon}}_\omega }({\mathbf{x}_t},t,\mathbf{y})$.
JCDI applies the diffusion and reverse processes to the model parameter space and uses the measured power trajectories as conditions to guide the generation process. 
Therefore, the joint probability of the reverse process becomes ${p_\omega }\left( {{\mathbf{x}_{0:T}}}|\mathbf{y} \right)$, allowing it to learn a mapping from observed system behaviors to the underlying model parameters that generate them.

{
\color{cyan}
\subsection*{Network architecture}
}
We introduce the Inverse Grid Transformer (IGT), a denoising neural network architecture based on the transformer encoder (illustrated in Fig.~\ref{fig:JCDI}). 
Originally developed for natural language processing, the Transformer has revolutionized the field while achieving remarkable success across multiple domains.
It converts the input, such as text, into numerical representations (tokens) and captures long-range dependencies among different tokens based on the attention mechanism~\citep{vaswani2017attention}. 

In our architecture, the transformer encoder receives three types of inputs:
(1) model parameters that are being denoised, (2) encoded power trajectories, and (3) encoded diffusion time step $t$. 
We do not add positional encoding here as the inputs have no natural distance measure between them.
Each input is tokenized with a separate linear encoder module (tokenizer).
The power trajectories act as the conditions. Before being converted into transformer tokens, they are processed through the trajectory encoders. The trajectory encoders employ ResNet architectures and extract latent features from the raw two-dimensional power trajectories~\citep{he2016deep}.
The diffusion step $t$ is sinusoidally embedded and incorporated as a part of inputs to provide information about noise level. 

The transformer encoder is composed of a stack of layers, each containing a multi-head attention layer followed by a feed forward network. Layer normalization is used within each sub-layer. 
The attention model calculates the \textcolor {cyan}{attention weights of input elements, which quantify their dependencies,} then produces the output representation through weighted summation of the inputs.
Specifically, each input token is transformed into three variables, query $Q$, key $K$, and value $V$, by linear projections in equation~\eqref{WQKV}. 
The model computes the attention scores by taking the dot product of the queries $Q$ and keys $K$, which measures the similarity between them.
The attention weights are then obtained by applying a softmax operation to the attention scores, ensuring that the attention weights sum up to 1 for each query. 
The model's output is an attention-weighted sum of values $V$ as expressed by equation~\eqref{attention}.
On this basis, the multi-head attention is implemented by performing several attention functions with different linear projections in parallel then concatenating the outputs.  
Based on the multi-head attention mechanism, the denoising neural network seeks to represent the \textcolor {cyan}{dependencies} among the CLM parameters, as well as the relationships between model parameters and conditioned power trajectories, enhancing parameter estimation performance. 
\begin{equation}
Q = {W_Q} \cdot X,{\rm{ }}K = {W_K} \cdot X,{\rm{ }}V = {W_V} \cdot X \label{WQKV},
\end{equation}
\begin{equation}
{\rm{Attention}}\left( {Q,K,V} \right) = {\rm{softmax}}\left( {\frac{{Q{K^T}}}{{\sqrt {{d_k}} }}} \right) \cdot V \label{attention},
\end{equation}
\noindent where ${W_Q}$, ${W_K}$, and ${W_V}$ are the transformation matrices; $X$ represents the input tokens; and ${{d_k}}$ denotes the dimension of keys.


{
\color{cyan}
\subsection*{Multi-event joint conditioning}
}
Due to parameter non-uniqueness, the traditional parameter estimation methods using a single disturbance may deduce various combinations of parameters, while some are not generalizable to other disturbances. Here, we address this challenge using the multi-event joint conditioning mechanism, where power trajectories under different fault-induced disturbances simultaneously serve as the condition inputs to deduce model parameters. 
Therefore, the posterior distribution of model parameters will be the conditional probability given multiple trajectory observations, as expressed by equation~\eqref{joint-condition}.
This serves to reduce parameter estimation uncertainties, thereby producing more robust and generalizable solutions.
\begin{equation}
		{P\left( {\bm{\mathrm{\theta}} |{\mathbf{y}_{1,0}},{\mathbf{y}_{2,0}}, \cdots ,{\mathbf{y}_{N,0}}} \right) = \frac{{P\left( {{\mathbf{y}_{1,0}},{\mathbf{y}_{2,0}}, \cdots ,{\mathbf{y}_{N,0}}|\bm{\mathrm{\theta}} } \right) \cdot P\left( \bm{\mathrm{\theta}}  \right)}}{{P\left( {{\mathbf{y}_{1,0}},{\mathbf{y}_{2,0}}, \cdots ,{\mathbf{y}_{N,0}}} \right)}}}   
		\label{joint-condition},
\end{equation}
\noindent where ${\mathbf{y}_{i,0}}$, $i$ = 1:$N$ denotes the measurement of power trajectories under the $i$-th fault event.

{
\color{cyan}
\subsection*{Training and inference procedures}
}

In the training stage, we randomly sample diffusion step $t \sim {\rm{Uniform}}\left( {\left\{ {1, \cdots ,T} \right\}} \right)$ and Gaussian noise $\bm{\mathrm{\varepsilon}} \sim \mathcal{N}\left( {\mathbf{0},\mathbf{I}} \right)$ then transform the original model parameters ${\mathbf{x}_0}$ into the noised latent ${\mathbf{x}_t}$ expressed by equation~\eqref{noised-x}.
\begin{equation}
{\mathbf{x}_t} = \sqrt {{{\bar \alpha }_t}} {\mathbf{x}_0} + \sqrt {1 - {{\bar \alpha }_t}} \bm{\mathrm{\varepsilon}} \label{noised-x},
\end{equation}
where ${\alpha _t} = 1 - {\beta _t}$, ${{\bar \alpha }_t} = \prod\nolimits_{s = 1}^t {{\alpha _s}} $. ${\beta _t}$ represents the variance schedule of the noising process.

The noise latent ${\mathbf{x}_t}$, diffusion step $t$, and conditioned power trajectories $\mathbf{y}$ are simultaneously input into the neural network IGT, which predicts the Gaussian noise as ${\bm{\mathrm{\varepsilon}} _\omega }\left( {{\mathbf{x}_t},t,\mathbf{y}} \right)$. Then, the denoising neural network is trained by minimizing the mean \textcolor {cyan}{square} error between the predicted and actually added Gaussian noises, which is calculated by equation~\eqref{loss}.
{
\color{cyan}
\begin{equation}
\min {\mathbb{E}_{\bm{\mathrm{\varepsilon}} ,{{\mathbf{x}}_0},t,\mathbf{y}}}{\left\| {\bm{\mathrm{\varepsilon}}  - {\bm{\mathrm{\varepsilon}} _\omega }\left( {\sqrt {{{\bar \alpha }_t}} {{\mathbf{x}}_0} + \sqrt {1 - {{\bar \alpha }_t}} \bm{\mathrm{\varepsilon}} ,t,{\mathbf{y}}} \right)} \right\|^2}\label{loss}.
\end{equation}
}

After completing the training process, we use the well-trained model to infer model parameters given the desired observations. In this inference stage, we start with randomly sampling ${\mathbf{x}_T} \sim \mathcal{N}\left( {\mathbf{0},\mathbf{I}} \right)$. At each step $t = T, \cdots ,1$, we use the desired power trajectories $\mathbf{y_0}$ as the condition to calculate the added noise with the trained neural network. 
Then, we substract it from the latent variable ${\mathbf{x}_t}$ and compute ${\mathbf{x}_{t - 1}}$ with equation~\eqref{denoise}. Therefore, the latent variables can be gradually denoised, and the noiseless estimation of model parameters ${\mathbf{x}_0}$ are obtained at the final step. Because of the probabilistic nature of the diffusion and reverse processes, we will deduce different parameter estimation solutions when repeatedly running the diffusion model, acquiring the posterior distribution of the predicted model parameters. 

\begin{equation}
{\mathbf{x}_{t - 1}} = \frac{1}{{\sqrt {{\alpha _t}} }}\left( {{\mathbf{x}_t} - \frac{{1 - {\alpha _t}}}{{\sqrt {1 - {{\bar \alpha }_t}} }}{\bm{\mathrm{\varepsilon}} _\omega }\left( {{\mathbf{x}_t},t,{{\mathbf{y}_0}}} \right)} \right) + {\sigma _t}\mathbf{z},
\label{denoise}
\end{equation}
where ${\sigma _t }={\sqrt {{\beta _t}} }$, $\mathbf{z} \sim \mathcal{N}\left( {\mathbf{0},\mathbf{I}} \right)$ \textcolor {cyan}{for $t > 1$, and ${\mathbf{z}} = {\mathbf{0}}$ for $t = 1$}.

\backmatter






\section*{Data availability}
{
\color{cyan}
The datasets generated during the study have been deposited in the Zenodo Repository (DOI:~\href{https://doi.org/10.5281/zenodo.18980716}{10.5281/zenodo.18980716})~\citep{zhu_2026_18980716}. They include: datasets for training and evaluation, trained model checkpoints, and source data for graphs and charts.
}

\section*{Code availability}
{
\color{cyan}
The source code for JCDI is available in a public repository at: \url{https://github.com/fq123fq/JCDI-power-grid}.
}

\bibliography{references}

{
\color{cyan}
\section*{Acknowledgments}
This work was supported by the Advanced Grid Modeling Program, Office of Electricity of the U.S. Department of Energy under Agreement 39917.

\section*{Author Information }
These authors contributed equally: Feiqin Zhu and Dmitrii Torbunov.

\subsection*{Authors and Affiliations}

\textbf{Feiqin Zhu}, Interdisciplinary Science Department, Brookhaven National Laboratory, Upton, 11973-5000, NY, USA. (Present address: School of Rail Transportation, Soochow University, Suzhou, 215131, China.)

\noindent \textbf{Dmitrii Torbunov}, Computing and Data Sciences Directorate, Brookhaven National Laboratory, Upton, 11973-5000, NY, USA.

\noindent \textbf{Zhongjing Jiang}, Environmental Science and Technologies Department, Brookhaven National Laboratory, Upton, 11973-5000, NY, USA. (Present address: Institute for Sustainability, Energy, and Environment, University of Illinois Urbana-Champaign, Urbana, 61801, IL, USA.)

\noindent \textbf{Tianqiao Zhao}, Interdisciplinary Science Department, Brookhaven National Laboratory, Upton, 11973-5000, NY, USA. (Present address: Department of Electrical Engineering, University of Texas at Arlington, Arlington, 76019, TX, USA.)

\noindent \textbf{Amirthagunaraj Yogarathnam}, Interdisciplinary Science Department, Brookhaven National Laboratory, Upton, 11973-5000, NY, USA.

\noindent \textbf{Yihui Ren}, Computing and Data Sciences Directorate, Brookhaven National Laboratory, Upton, 11973-5000, NY, USA.

\noindent \textbf{Meng Yue}, Interdisciplinary Science Department, Brookhaven National Laboratory, Upton, 11973-5000, NY, USA.

\subsection*{Author contributions} 
Y.R., M.Y., F.Z., and D.T. conceptualized this work. 
D.T. and F.Z. developed the machine learning models.
F.Z., T.Z., and A.Y. conducted power load simulation and analysis.  
F.Z. and Z.J. conducted sensitivity and uncertainty analysis.
F.Z. and D.T. drafted the manuscript.
D.T., Y.R., M.Y., and Z.J. reviewed and edited the manuscript.
Y.R. and M.Y. supervised this work.
M.Y. provided funding support.
All authors contributed to the discussions.

\subsection*{Corresponding Authors}
Correspondence to Yihui Ren.

\section*{Ethics Declarations}
\subsection*{Competing interests}
The authors declare no competing interests.

\section*{Additional information}
\subsection*{Supplementary Information}
This work contains supplementary material available at: \url{https://doi.org/10.1038/s44172-026-00670-z}.

}


\end{document}


\title[Article Title]{Supplementary Information for \\\\\textbf{Diffusion Model-based Parameter Estimation in Dynamic Power Systems}}

\author[1,4]{\fnm{Feiqin} \sur{Zhu}}\email{fqzhu@suda.edu.cn}  
\equalcont{These authors contributed equally.}

\author[2]{\fnm{Dmitrii} \sur{Torbunov}}\email{dtorbunov@bnl.gov}
\equalcont{These authors contributed equally.}

\author[3,5]{\fnm{Zhongjing} \sur{Jiang}}\email{zjiang35@illinois.edu}

\author[1,6]{\fnm{Tianqiao} \sur{Zhao}}\email{tianqiao.zhao@uta.edu}

\author[1]{\fnm{Amirthagunaraj} \sur{Yogarathnam}}\email{rajyogar@ieee.org}

\author*[2]{\fnm{Yihui} \sur{Ren}}\email{yren@bnl.gov}  

\author[1]{\fnm{Meng} \sur{Yue}}\email{yuemeng@bnl.gov}

\affil[1]{\orgdiv{Interdisciplinary Science Department}, \orgname{Brookhaven National Laboratory}, \orgaddress{\city{Upton}, \postcode{11973-5000}, \state{NY}, \country{USA}}}


\affil[2]{\orgdiv{Computing and Data Sciences Directorate}, \orgname{Brookhaven National Laboratory}, \orgaddress{\city{Upton}, \postcode{11973-5000}, \state{NY}, \country{USA}}}


\affil[3]{\orgdiv{Environmental Science and Technologies Department}, \orgname{Brookhaven National Laboratory}, \orgaddress{ \city{Upton}, \postcode{11973-5000}, \state{NY}, \country{USA}}}

\affil[4]{Present address: \orgdiv{School of Rail Transportation}, \orgname{Soochow University}, \orgaddress{ \city{Suzhou}, \postcode{215131}, \country{China}}}

\affil[5]{Present address: \orgdiv{Institute for Sustainability, Energy, and Environment}, \orgname{University of Illinois Urbana-Champaign}, \orgaddress{ \city{Urbana}, \postcode{61801}, \state{IL}, \country{USA}}}

\affil[6]{Present address: \orgdiv{Department of Electrical Engineering}, \orgname{University of Texas at Arlington}, \orgaddress{ \city{Arlington}, \postcode{76019}, \state{TX}, \country{USA}}}

\maketitle

\newpage

\hypersetup{linkcolor=black}
\tableofcontents
\hypersetup{linkcolor=blue}

\newpage



\renewcommand{\theequation}{Supplementary Equation \arabic{equation}}
\renewcommand{\figurename}{Supplementary Fig.}

\renewcommand{\tablename}{Supplementary Table.}

\section{CMPLDWG compositions and characteristics}
\label{sec:Appendix-A}

To address the rapid evolution of electric power loads, Western Electricity Coordinating Council has developed \textcolor {cyan}{CMPLDWG, an advanced composite load model (CLM) with distributed generation}~\citep{CMPLDW-DG}. 
It consists of three, three-phase induction motors with different characteristics (motors A, B, and C), one single-phase induction motor (motor D), one power electronic load, one static load, and one distributed energy resource (DER)~\citep{WECC}. This allows the model to represent different electric characteristics of power loads and flexibly change the fractions of each composition according to actual power loads.

\begin{figure}[htbp]
	\centering
    \includegraphics[width=8cm]{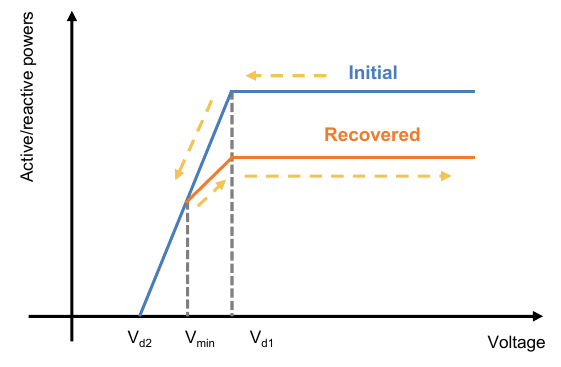} 
	\caption{\textbf{Power changes of power electronic load when tripping.} The blue line illustrates its powers during initial operation and the occurrence of tripping, while the orange line represents power changes throughout the recovery process. The yellow arrows indicate the state transition process.}
	\label{fig:trip}
\end{figure}

\begin{itemize}[leftmargin=*]
\item Motor A characterizes the three-phase induction motors that drive low-inertia constant torque loads. Typical examples include positive displacement compressors and pumps~\citep{EPRI}. Motor B represents the high-inertia variable torque loads, such as large fans, and motor C is used to model the low-inertia variable torque loads, such as centrifugal pumps. They are modeled by fifth-order differential-algebraic equations based on electromagnetic and electromechanical equations with different parameters~\citep{ma2020mathematical}. 
\item Motor D represents the single-phase induction motor, e.g., residential air conditioners and heat ventilation. In CMPLDWG, it is developed as a ``performance model'' based on laboratory tests. 
\item Static loads are characterized by the ZIP model, which consists of constant impedance (Z), constant current (I), and constant power (P) components.
\item The power electronic load represents an aggregation of inverter-interfaced or electronic coupled loads, such as consumer electronic devices (e.g., computers). Supplementary Figure~\ref{fig:trip} depicts the power-voltage relationship of electronic load.
\item DER Version A (DER\_A) is newly developed in load modeling to represent the aggregation of inverter-based generation (e.g., photovoltaic)~\citep{CMPLDW-DG}. 
Based on its block diagram and specification~\citep{DER-A}, DER\_A enables different control modes and their flexible switching, including constant Q-control and constant power factor control. It also contains various current-limit modes, including P-priority and Q-priority modes. In our work, we select the conventional modes: constant Q-control and P-priority mode. The nonlinearities of DER mainly include the current and power limits, the dead-bands, and tripping behavior when satisfying the tripping conditions. 
\end{itemize}

\bmhead{Stalling behavior of a single-phase induction motor}

\begin{figure}[htbp]
	\centering
    \includegraphics[width=0.95\textwidth]{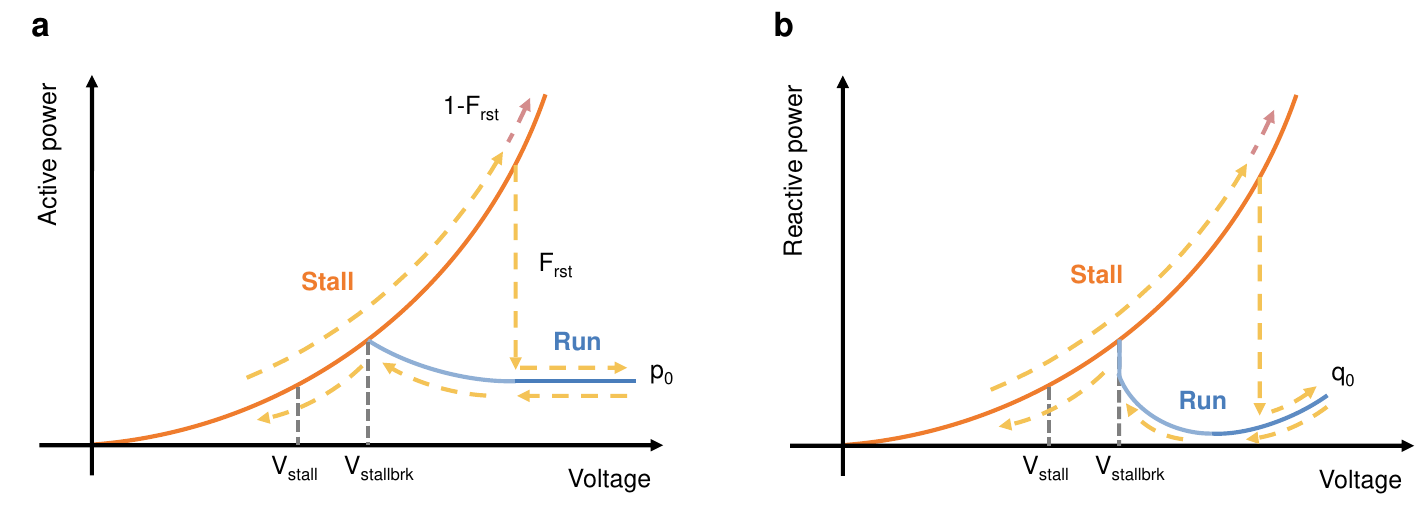} 
	\caption{\textbf{Motor D power changes when stalling.} The blue line illustrates motor D operating in run state, while the orange line represents its power change in stall state. The yellow arrows indicate the state transition process. \textbf{a} active power and \textbf{b} reactive power.}
	\label{fig:stall}
\end{figure}

Single-phase motors are prone to stalling when compared with three-phase motors due to their low torque characteristics at low voltage levels. There is insufficient motor torque to overcome the load torque. Therefore, the motor stops. Supplementary Figure~\ref{fig:stall} shows the power change of motor D during the stalling and recovering processes.

When the terminal voltage drops below the motor stalling voltage ${V_{stall}}$ for a time duration of ${T_{stall}}$, motor D transitions from the \textit{run} state to \textit{stall} state. The active and reactive powers of motor D are proportional to the square of voltage in stall state, expressed by~\eqref{stall-MD}, leading to high power consumption.
\begin{equation}
\begin{array}{l}
p = {V^2}/{R_{stall}}\\
q =  - {V^2}/{X_{stall}}
\end{array} \label{stall-MD},
\end{equation}
\noindent where ${R_{stall}}$ and ${X_{stall}}$ respectively denote the stall resistance and reactance.

When the terminal voltage recovers and becomes higher than the restarting voltage ${V_{rst}}$ of the stalled motors for a time duration of ${T_{rst}}$, a portion (${F_{rst}}$) of the motor D loads transition to run state, while the rest remains in stall state.

\bmhead{Tripping behavior of power electronic load} 
As shown in Supplementary Figure~\ref{fig:trip}, when the terminal voltage maintains above ${V_{d1}}$, the electronic load consumes constant active and reactive power.
However, when the voltage drops below the ${V_{d1}}$, the powers reduce linearly, indicating undervoltage trip of the power electronic loads.
The powers become zero when the voltage reduces to ${V_{d2}}$.
As the voltage recovers, there will be a certain fraction of loads reconnecting to the system. Thus, the active and reactive powers increase and stabilize at values lower than their initial levels, which is calculated by~\eqref{el-rec}. 

\begin{equation}
\begin{array}{l}
Fv{l_{rec}} = \frac{{\left( {{V_{\min }} - {V_{d2}}} \right) + frcel \cdot \left( {{V_{d1}} - {V_{\min }}} \right)}}{{{V_{d1}} - {V_{d2}}}}\\
{p_{el,rec}} = {p_{el,0}} \cdot Fv{l_{rec}}\\
{q_{el,rec}} = {q_{el,0}} \cdot Fv{l_{rec}}
\end{array} \label{el-rec},
\end{equation}
\noindent where ${{V_{\min }}}$ is the minimum value of the terminal voltage during the tripping process, $frcel$ is the fraction of electronic loads that reconnect to the system after voltage recovery, and $Fv{l_{rec}}$ denotes the proportion of recovered powers. ${p_{el,0}}$ and ${q_{el,0}}$ respectively represent the initial active and reactive powers. ${p_{el,rec}}$ and ${q_{el,rec}}$ designate the recovered powers.

\section{CMPLDWG dynamic simulation}
\label{sec:Appendix-simulation}
To implement dynamic simulation, CMPLDWG is supplied by the bulk transmission system as depicted in Supplementary Figure~\ref{fig:T-L-sim}.
The transmission system represents an IEEE 39-bus system~\citep{bhui2016real} established in ePHASORSIM. 
The voltage profiles at Bus 9, the transmission system's point of interconnection (POI), are fed into CMPLDWG. 

We simulate three-phase-to-ground electrical faults in the transmission system.
Three separate fault events are implemented, each with a fault impedance of $1 \times {10^{ - 5}} \Omega $ at different bus locations~\citep{ePHASORSIM}.
The corresponding fault clearing times and locations are detailed in Supplementary Table~\ref{tab:list-fault}. 
The fault clearing times are selected from a relatively wide yet realistic range, ensuring our analysis encompasses a comprehensive set of potential grid conditions. Based on the empirical value of three to six cycles, the fault clearing time also can be influenced by uncertain delays, voltage level of the transmission systems, and advancement of modern protection technologies~\citep{Kundur,Modern-protect}. 
\begin{figure}[htbp]
	\centering
    \includegraphics[width=0.85\textwidth]{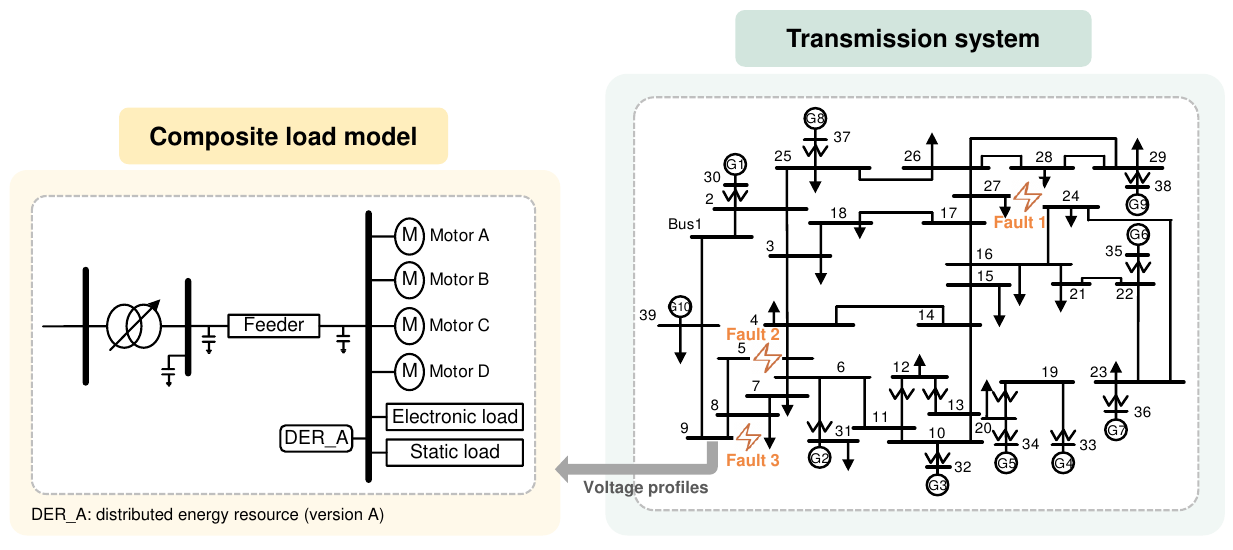} 
	\caption{
\textbf{Schematic diagram of CMPLDWG dynamic simulation.}
 Load model represented by CMPLDWG is supplied by the IEEE 39-bus transmission system. Electrical faults are simulated at different bus locations in the transmission system.
    }
	\label{fig:T-L-sim}
\end{figure}

\begin{table}[htbp]
	\caption{List of electrical fault events}
	\centering
	\begin{tabular}{llll}
		\toprule
		Fault index     & Bus location    & Clearing time & Dynamic characteristics  \\
		\midrule
		No. 1           &  27          &  135 ms & Ordinary \\
  	  No. 2           &  5           &  135 ms & Power electronic load tripping \\
            No. 3           &  9           &  44 ms &  Motor D stalling \\
		\bottomrule
	\end{tabular}
	\label{tab:list-fault}
\end{table}

We use the voltage profiles under electric fault disturbances as the inputs and compare the CMPLDWG's power responses with different parameter estimates. 
This approach is well established as evidenced by its application to evaluate the dynamic responses of load models and DERs~\citep{ma2020mathematical,DER-A}.

\begin{figure}[htbp]
	\centering
    \includegraphics[width=0.9\textwidth]{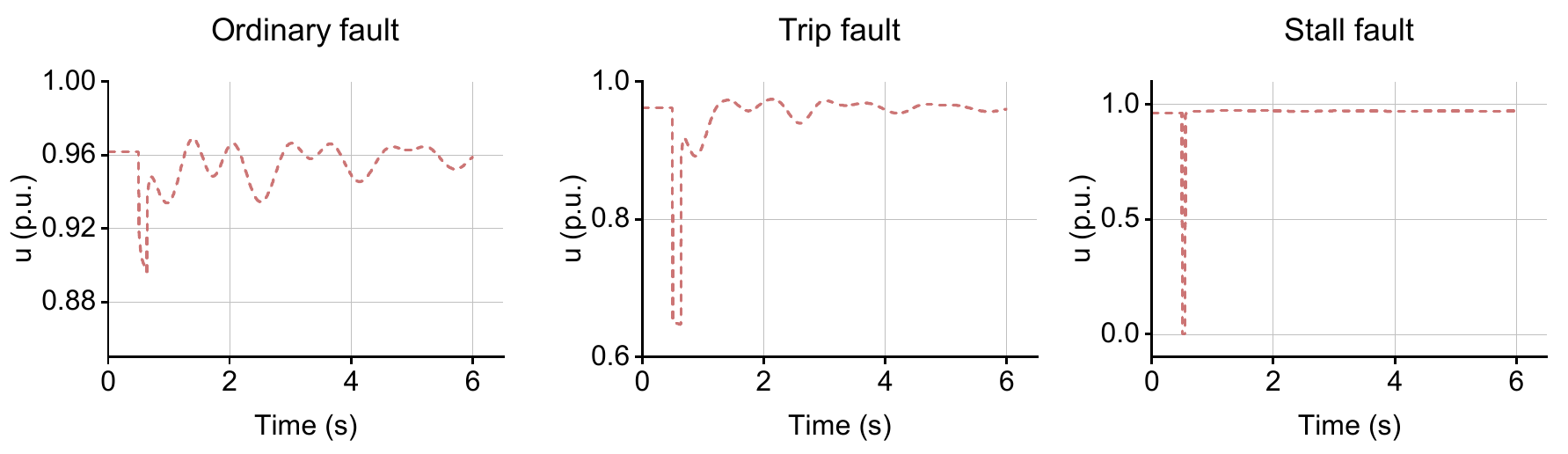} 
	\caption{
\textbf{Voltage trajectories under different fault disturbances.}
Voltage profiles are measured at the POI. They exhibit distinct voltage drops and recovery processes when electrical faults occur at different bus locations.
    }
	\label{fig:trajs-u}
\end{figure}

Supplementary Figure~\ref{fig:trajs-u} depicts the voltage profiles under the three electric fault events. 
The first electric fault (ordinary fault) occurs on Bus 27, which is far from the POI. Therefore, this fault disturbance induces a minor effect, where the voltage reduces slightly during the fault (minimum to 0.90 p.u.).
The second electric fault (trip fault) occurs on Bus 5, which is much closer to the POI and load model. 
As a result, the voltage drops to 0.65 p.u., which is below the voltage tripping threshold (${V_{d1}}$=0.8 p.u.) for the power electronic load in CMPLDWG (Supplementary Fig.~\ref{fig:trip})~\citep{EPRI}. This causes the power electronic load to trip during the fault event.
The third electric fault (stall fault) is located on the POI. The voltage becomes zero during the fault interval (44 ms). This induces the stalling behavior of motor D, where the voltage drops lower than ${V_{stall}}$ for a time duration of ${T_{stall}}$~\citep{EPRI}. In our study, ${V_{stall}}=0.6$ p.u., and ${T_{stall}}=30$ ms.

\section{Parameter sensitivity analysis}
\label{sec:Appendix-B}

In this work, sensitivity analysis based on Sobol's method~\citep{SOBOL2001271,tosin2020tutorial,SALTELLI2002280,Iwanaga2022,Herman2017} is performed for parameter reduction. Sobol's method is a variance-based global sensitivity analysis method, which decomposes the variance of output into contributions from individual input parameters and their interactions. Therein, first-order index measures the individual effect of the input variable on the output. Second-order index indicates the effect of interaction between different input variables. The total-effect index calculates the full contribution of the input variable on the output variance, which is the sum of the indices of different orders.

\begin{figure}[htbp]
	\centering
    \includegraphics[width=0.95\textwidth]{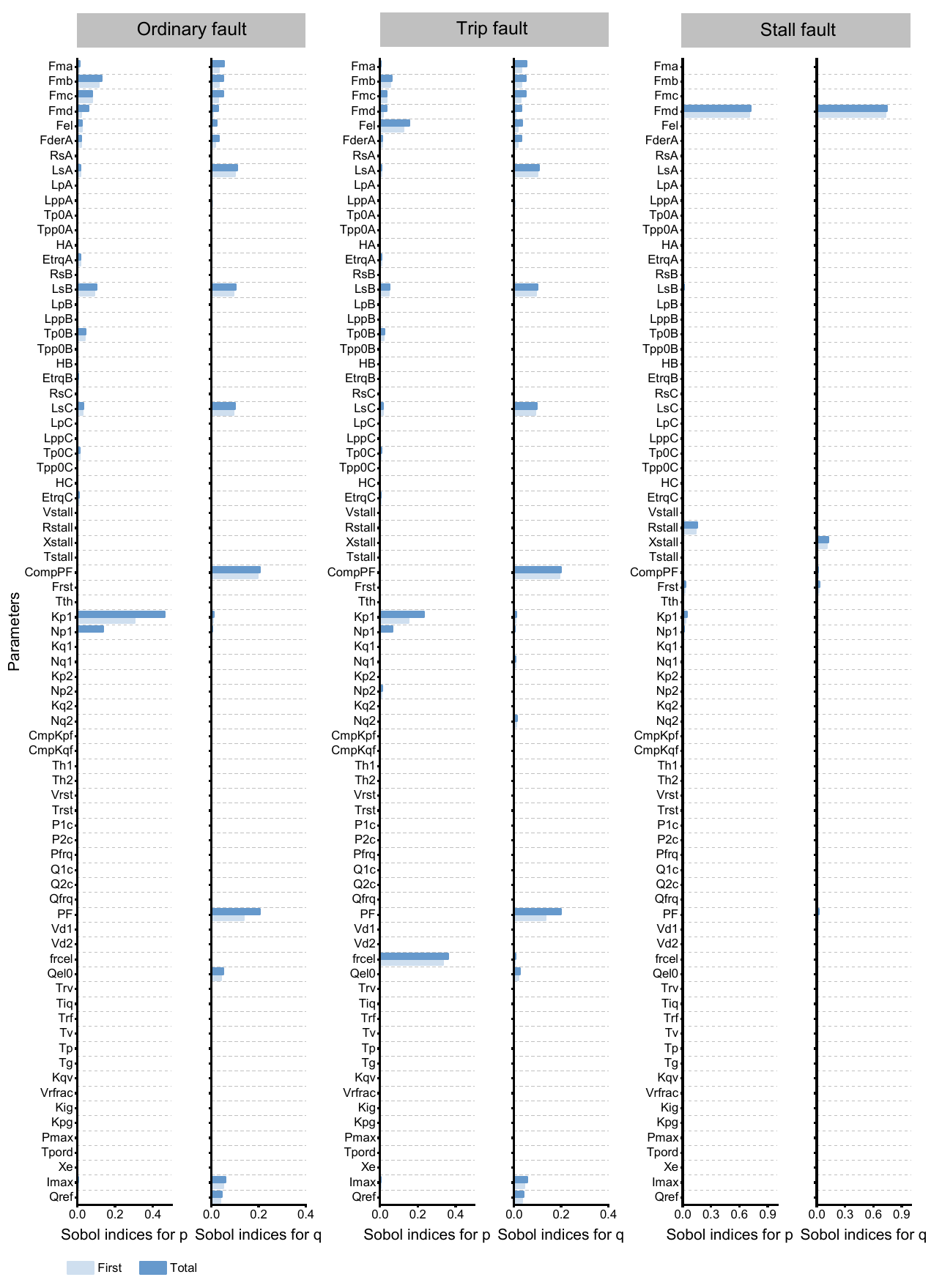} 
	\caption{\textbf{Sobol's indices of model parameters.} The Sobol indices of model parameters in CMPLDWG are calculated for both active and reactive power trajectories under different fault events. The bars in light blue and dark blue correspondingly represent the first-order and total-effect indices: \textbf{(left)} ordinary fault, \textbf{(middle)} trip fault, and \textbf{(right)} stall fault.}
	\label{fig:sobol-detail}
\end{figure}

\begin{figure}[htbp]
	\centering
    \includegraphics[width=0.95\textwidth]{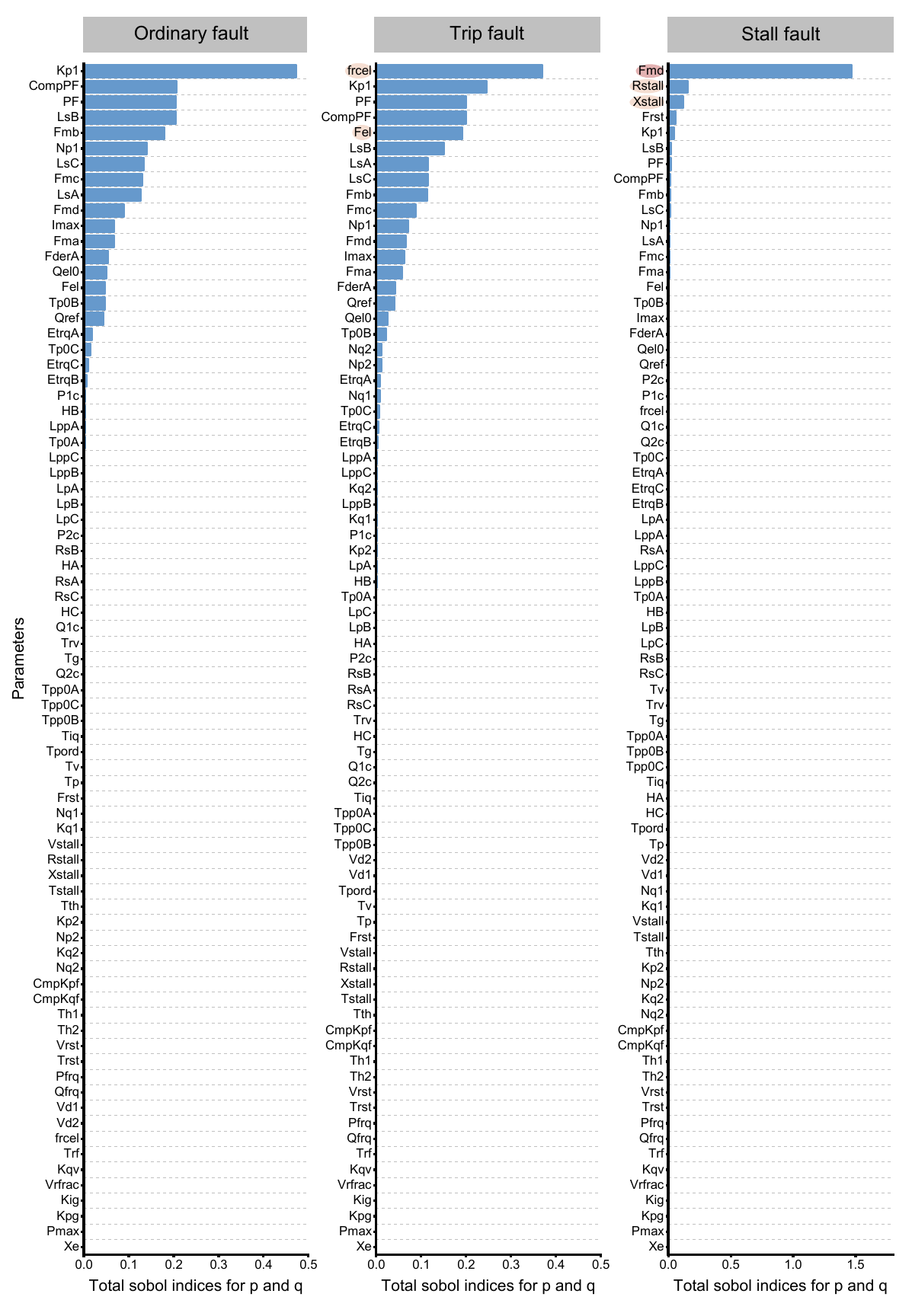} 
	\caption{\textbf{The rankings of parameter sensitivity under different fault events.} The model parameters in CMPLDWG are ranked according to the total-effect indices for both active and reactive power trajectories. \textcolor {cyan}{The orange and pink ellipses highlight the parameters exhibiting markedly enhanced sensitivity.} \textbf{(left)} ordinary fault, \textbf{(middle)} trip fault, and \textbf{(right)} stall fault.}
	\label{fig:sobol-ranking}
\end{figure}

To gain insight into the influences of model parameters on the system dynamic behavior under different fault-induced disturbances, we calculate the Sobol indices of CMPLDWG parameters for active and reactive power trajectories under the ordinary, trip, and stall faults (Supplementary Fig.~\ref{fig:sobol-detail}).
Then, the model parameters are ranked according to their total-effect indices (Supplementary Fig.~\ref{fig:sobol-ranking}). 
Most of the parameters present consistent sensitivity under different fault events. However, $frcel$ and ${F_{el}}$ become much more sensitive when there are power electronic loads tripping. With the occurrence of motor D stalling, ${F_{md}}$ emerges as the most sensitive parameter, exhibiting substantially higher sensitivity compared to other parameters. Additionally, the sensitivities of ${R_{stall}}$, ${X_{stall}}$, and ${F_{rst}}$ increase markedly. Comprehensively considering the sensitivity rankings under the three fault events, 30 sensitive parameters are selected for parameter estimation, which are listed in Supplementary Table~\ref{tab:sensitive-paras}.

\begin{table}[htbp]
	\caption{List of selected model parameters for identification}
	\centering
	\begin{tabular}{llll}
		\toprule
		Notations     & Description      & \multirow{2}{*}{\makecell[l]{ Sampling\\ ranges}}    & \multirow{2}{*}{\makecell[l]{ Default\\ values}} \\\\
		\midrule
		${F_{ma}}$     & Motor A fraction               & [0.1,0.3]   &  0.2 \\
		${F_{mb}}$     & Motor B fraction               & [0.1,0.3]   &  0.2  \\
		${F_{mc}}$     & Motor C fraction               & [0.1,0.3]   &  0.2 \\
  	${F_{md}}$     & Motor D fraction               & [0.1,0.3]   &   0.2  \\
  	${F_{el}}$     & Electronic load fraction       & [0.1,0.3]   & 0.2 \\
  	${F_{derA}}$    & DER\_A fraction         & [-0.3,-0.1]   & -0.2 \\
  	${L_{sA}}$     & Synchronous reactance (pu) of motor A      & [1.5,3]  & 1.8 \\
    ${E_{trqA}}$     &  \multirow{2}{*}{\makecell[l]{ Speed exponent for mechanical toque\\ of motor A}}  & [0,1]  & 0 \\\\
  	${L_{sB}}$     & Synchronous reactance (pu) of motor B       & [1.5,3] &  1.8 \\
  	${T_{p0B}}$     & \multirow{2}{*}{\makecell[l]{Transient open circuit time constant (sec.)\\ of motor B}} & [0.08,0.12] & 0.1  \\\\
    ${E_{trqB}}$     &  \multirow{2}{*}{\makecell[l]{ Speed exponent for mechanical toque\\ of motor B}}  & [1.5,2.5]  & 2 \\\\
  	${L_{sC}}$     & Synchronous reactance (pu) of motor C       & [1.5,3] & 1.8 \\
  	${T_{p0C}}$     & \multirow{2}{*}{\makecell[l]{ Transient open circuit time constant (sec.)\\ of motor C}}   &  [0.08,0.12] & 0.1 \\\\
  	${E_{trqC}}$     & \multirow{2}{*}{\makecell[l]{ Speed exponent for mechanical toque\\ of motor C}}       &   [1.5,2.5]  & 2  \\\\
  	${R_{stall}}$     & Stall resistance (pu) of motor D       & [0.08,0.12]& 0.1 \\
  	${X_{stall}}$     & Stall reactance (pu) of motor D       & [0.08,0.12]& 0.1 \\
  	${CompPF}$     & Power factor of motor D        & [0.9,1] & 0.98  \\
  	${F_{rst}}$     &  \multirow{2}{*}{\makecell[l]{ Fraction of load that can restart after stalling\\ of motor D }}      &  [0.15,0.3] & 0.2  \\\\
  	${K_{p1}}$     & Active power coefficient of motor D in run state      & [-1,1] & 0 \\
  	${N_{p1}}$     & Active power exponent of motor D in run state      & [0.5,1.5] & 1 \\
  	${N_{q1}}$     & Reactive power exponent of motor D in run state     & [1,3]  & 2  \\
  	${N_{p2}}$     & Active power exponent of motor D in stall state      & [1.6,4.8] & 3.2 \\
  	${N_{q2}}$     & Reactive power exponent of motor D in stall state      & [1.25,3.75]  & 2.5  \\
      ${P_{1c}}$     &   Coefficient of  ${P_{1}}$ for static load    & [0.3,0.5]  & 0.4   \\
      ${P_{2c}}$     &   Coefficient of  ${P_{2}}$ for static load    & [0.5,0.7]  & 0.6   \\
  	${PF}$     & Power factor of static load     &  [0.9,1]  & 0.95 \\
  	${f_{rcel}}$     & \multirow{2}{*}{\makecell[l]{ Fraction of electronic load that recovers from\\ low voltage trip }}      &  [0.5,0.9] & 0.75 \\\\
  	${Q_{el0}}$     & Initial reactive power of power electronic load   & [0.1,0.3] & 0.2 \\
  	${I_{max}}$     & Maximum current of DER\_A (pu)       & [1,1.5]& 1.2 \\
  	${Q_{ref}}$     & Reactive power reference of DER\_A (pu)       & [0.1,0.3]  &  0.2 \\
		\bottomrule
	\end{tabular}
	\label{tab:sensitive-paras}
\end{table}

\section{Supplementary information on JCDI}
\label{sec:Appendix-C}

\subsection{Supplementary training procedures}
\label{sec:JCDI-training}
\bmhead{Data generation}
We randomly sample the sensitive model parameters from a uniform distribution within the ranges specified in Supplementary Table~\ref{tab:sensitive-paras}. Then, dynamic simulations described in Supplementary Section~\ref{sec:Appendix-simulation} are performed using these parameters in CMPLDWG to obtain the dynamic responses.
The sampling ranges are determined by considering both physical constraints of electric parameters (such as ${L_{sA}}$) and perturbations of up to 50\% to the settable parameters (such as ${F_{ma}}$). 
The total dataset size is 300000, including 250000 data samples for training and 50000 for testing. 

\bmhead{Model training}
Supplementary Table~\ref{tab:hyper-para-JCDiff} presents the Joint Conditional Diffusion Model-based Inverse Problem Solver (JCDI) hyperparameters. 
The trajectory encoder employs a ResNet architecture. It encodes power trajectories using a stem layer (kernel size 4, stride 4, 32 kernels) followed by two pairs consisting of a residual block and a downsampling layer. \textcolor {cyan}{For CDI, each downsampling layer has a stride of 4 and a kernel size of 4, containing 128 and 256 convolutional kernels, respectively. This architecture transforms the input power trajectory of shape (2, 512) into a latent representation of shape (256, 8).  
For JCDI, the model is conditioned on multiple fault events using separate trajectory encoders for each event. 
To manage the increase of tokens when processing multiple events, we increase the stride of the first downsampling layer to 8, producing a latent representation of shape (256, 4) per event. 
This introduces more aggressive downsampling; alternative approaches such as adaptive pooling may yield further performance improvements.}
The transformer encoder is composed of three layers, each containing a multi-head attention mechanism with four attention heads. 
For the diffusion process, the diffusion step is 200, and a linear variance schedule is used to add noise. The Adam optimizer is used to minimize the loss function with a learning rate of $1 \times {10^{ - 4}}$ and a batch size of 128. Training is implemented with PyTorch on a NVIDIA GeForce RTX 3090 graphics processing unit (GPU).
\begin{table}[htbp]
	\caption{Hyperparameters for JCDI}
	\centering
	\begin{tabular}{lllll}
		\toprule
		Parameter types&Hyperparameters     & Notations         & Values \\
		\midrule
		\multirow{3}{*}{Diffusion model} & Diffusion step     & $T$            &  200 \\
  	&Initial value of variance schedule     &  ${\beta _0}$      &  0.0001 \\
        &Final value of variance schedule     &  ${\beta _T}$      &  0.005 \\
        \midrule
        \multirow{2}{*}{Training} &Batch size     &  $B$      &  128  \\
        &Learning rate     &  $lr$      &  $1 \times {10^{ - 4}}$  \\
        \midrule
        \multirow{3}{*}{ \makecell[l]{Trajectory\\ encoder}   } & Stem layer (kernel size, number)    &  -     &  [$1\times4$, 32]  \\
        &Downsampling layer 1    &  -     &  [$1\times4$, 128]  \\
        &Downsampling layer 2    &  -     &  [$1\times4$, 256]  \\
        \midrule
        \multirow{4}{*}{ \makecell[l]{Transformer\\ encoder} } &Number of layers   &  -     &  3  \\
        &Number of heads  &  -     &  4  \\
        &Features in the input   &  -     &  256  \\
        &Features in the feed-forward network   &  -     &  512  \\
		\bottomrule
	\end{tabular}
	\label{tab:hyper-para-JCDiff}
\end{table}

\subsection{Supplementary results}
\label{sec:supp-JCDI-results}
\begin{figure}[htbp]
	\centering
    \includegraphics[width=9cm]{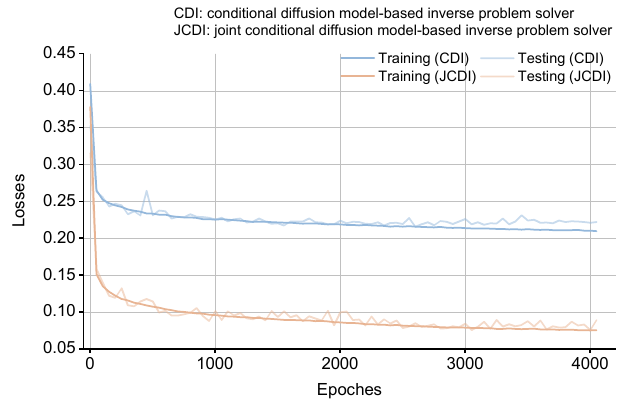} 
	\caption{
    \textbf{Loss evolution in the training progress.} 
    The respective dark and light blue lines represent the training and testing losses for CDI, while the orange lines are those for JCDI.}
	\label{fig:loss}
\end{figure}

\bmhead{Training progress} 
Supplementary Figure~\ref{fig:loss} shows the evolution of training and testing losses in the training progress. As the number of epochs increases, they both decrease smoothly and level off at similar values, demonstrating the models avoid overfitting on the training data. It is also important to note that the losses for JCDI are substantially lower than those for CDI (Conditional Diffusion Model-based Inverse Problem Solver).

\begin{figure}[htbp]
	\centering
     \includegraphics[width=0.98\textwidth]{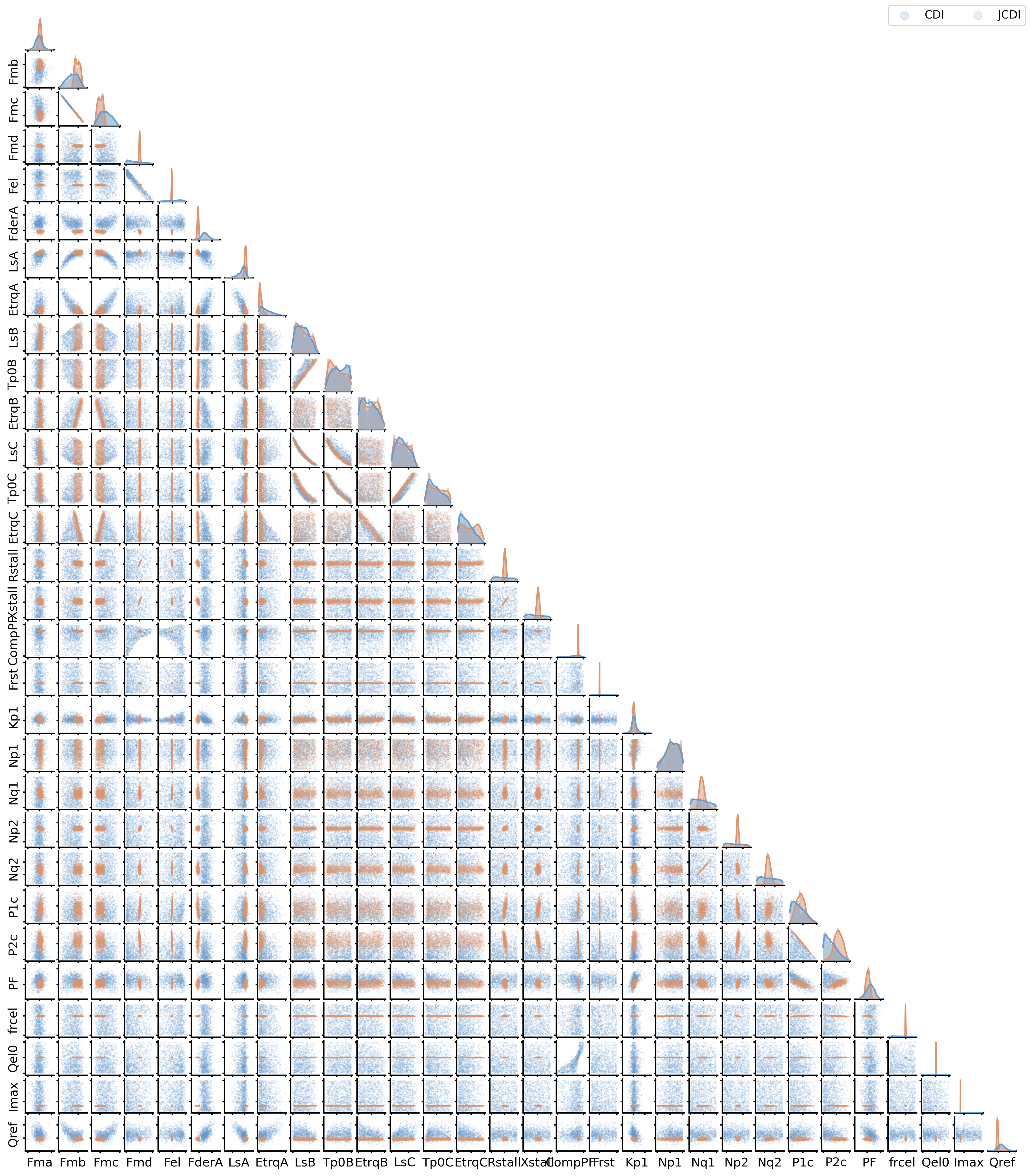} 
	\caption{\textbf{Posterior results of parameter estimation.} The subplots in the lower triangular region display the scatter plots of parameter estimates. The marginal posterior distribution for each model parameter is plotted along the diagonal. Blue plots represent CDI results, while orange plots show JCDI results. \textcolor {cyan}{The plots for the posterior results are constructed using 1000 parameter estimates.}}
	\label{fig:parameter-correlation}
\end{figure}

\bmhead{Parameter estimation uncertainties} 
Supplementary Figure~\ref{fig:parameter-correlation} provides the complete posterior results of parameter estimation.
The eigenvalue distributions of CMPLDWG with actual parameters and those estimated by JCDI are compared in Supplementary Figure~\ref{fig:Eigenvalues}.
\begin{figure}[htbp]
	\centering
    \includegraphics[width=0.8\textwidth]{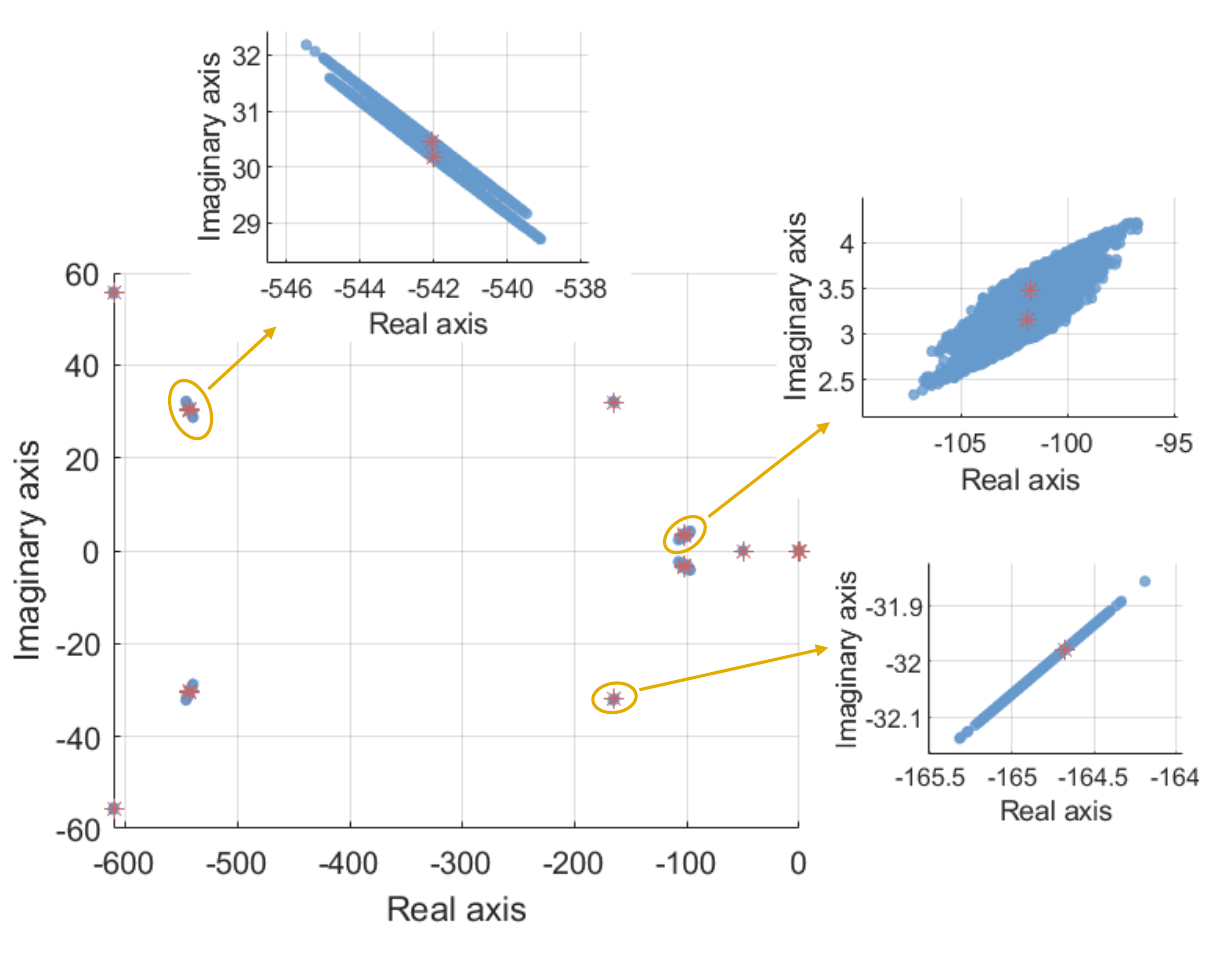} 
\caption{\textbf{Eigenvalue distributions of CMPLDWG.} 
    The blue dots represent the eigenvalues of CMPLDWG with \textcolor {cyan}{1000} parameter estimates deduced by JCDI, while the pink stars denote the actual system eigenvalues. \textcolor {cyan}{The yellow annotations indicate the enlarged eigenvalue distributions.}
    }
	\label{fig:Eigenvalues}
\end{figure}

\begin{figure}[htbp]
	\centering
     \includegraphics[width=0.98\textwidth]{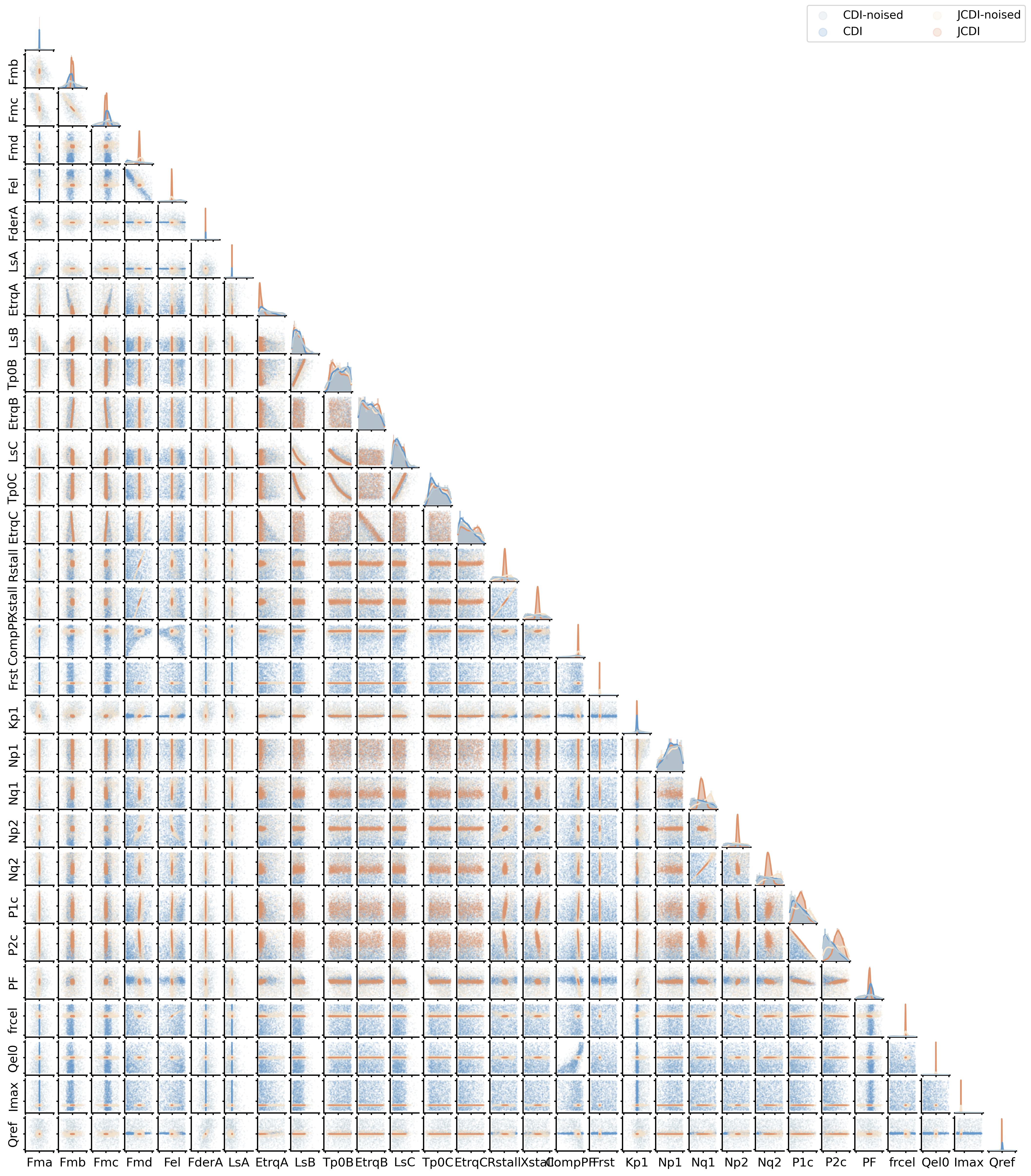} 
	\caption{\textbf{Posterior results of parameter estimation considering measurement noise at an SNR of 50 dB.}
    Parameters estimates of CDI and JCDI with noise-free measurements are shown in blue and orange, respectively, while the respective estimation results of CDI and JCDI with noisy measurements are presented in green and yellow.
    \textcolor {cyan}{The statistical plots are constructed using 1000 parameter estimates.}} 
	\label{fig:parameter-correlation-noise}
\end{figure}

\begin{figure}[htbp]
	\centering
    \includegraphics[width=0.95\textwidth]{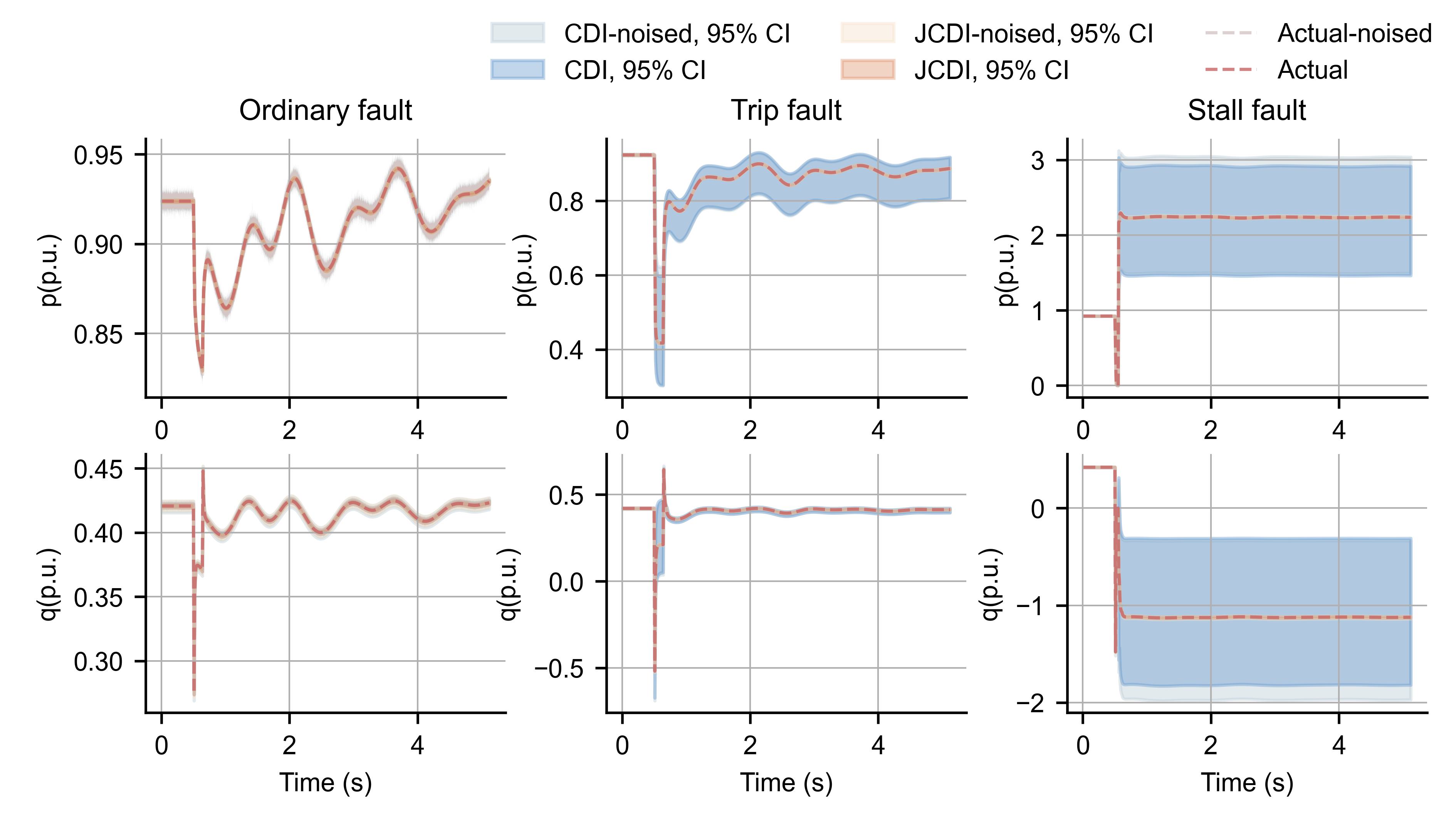} 
\caption{\textbf{Prediction results of dynamic responses considering measurement noise at an SNR of 50 dB.} 
    The pink and brown dashed lines are the actual power measurements before and after adding Gaussian noises. When inferred with noise-free measurements, the predicted power trajectories of CDI and JCDI with 95\% \textcolor {cyan}{credible intervals (CIs)} are displayed with blue and orange areas, respectively. The respective prediction results of CDI and JCDI with noisy measurements are presented by green and yellow areas. \textcolor {cyan}{The plots are constructed using 1000 data samples.}}
	\label{fig:traj-result-noisy}
\end{figure}

\bmhead{\textcolor {cyan}{Impact of measurement noise}}
When training CDI and JCDI with noisy data, we expand the training dataset size to 480000 to avoid overfitting.
Supplementary Figure~\ref{fig:parameter-correlation-noise} presents the complete posterior results of parameter estimation considering measurement noise at a 50 dB signal-to-noise ratio (SNR).
Supplementary Figure~\ref{fig:traj-result-noisy} depicts the prediction results of dynamic responses.

\begin{figure}[htbp]
	\centering
    \includegraphics[width=0.9\textwidth]{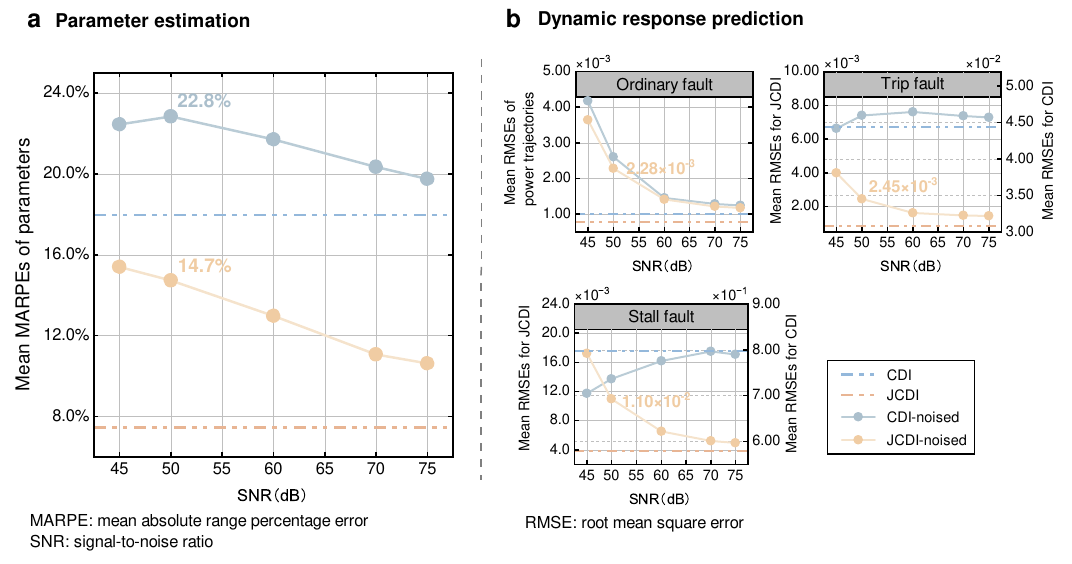} 
\caption{\textcolor {cyan}{\textbf{Model sensitivity to noise level.}} JCDI and CDI are retrained with noisy data at SNRs ranging from 45 to 75 dB. Then, they infer model parameters conditioned on power trajectories at different noise levels: \textbf{a} influence of noise level on parameter estimation accuracy and \textbf{b} influence of noise level on power trajectory prediction accuracy.
\textcolor {cyan}{Results for CDI and JCDI with noisy measurements are shown in green and yellow, respectively. Results for CDI and JCDI with noise-free measurements are respectively represented by the blue and orange dashed lines.}
}
	\label{fig:error-with-noise}
\end{figure}

\textcolor {cyan}{
We further investigate the sensitivity of our model to noise level.
Supplementary Figure~\ref{fig:error-with-noise} shows the parameter estimation and dynamic response prediction accuracy at different noise levels.
Within an SNR range of 45–75 dB, the mean absolute range percentage error (MARPE) of model parameters for JCDI maintains below 16\%.
For dynamic response prediction, the prediction accuracy based on JCDI also decreases as the SNR decreases under different fault disturbances.
It shows a similar downward trend under the ordinary fault for CDI, but it varies little around the noise-free prediction under the trip and stall faults as CDI is conditioned solely on the ordinary fault.
For SNRs ranging from 45 to 75 dB, the mean root mean square errors (RMSEs) of power trajectories for JCDI under the ordinary and trip faults remain below $4 \times {10^{ - 3}}$.
Under the stall fault, it increases modestly to $1.72 \times {10^{ - 2}}$ at an SNR of 45 dB then remains below $1.10 \times {10^{ - 2}}$ for SNRs exceeding 50 dB.
}

\begin{figure}[htbp]
	\centering
     \includegraphics[width=0.98\textwidth]{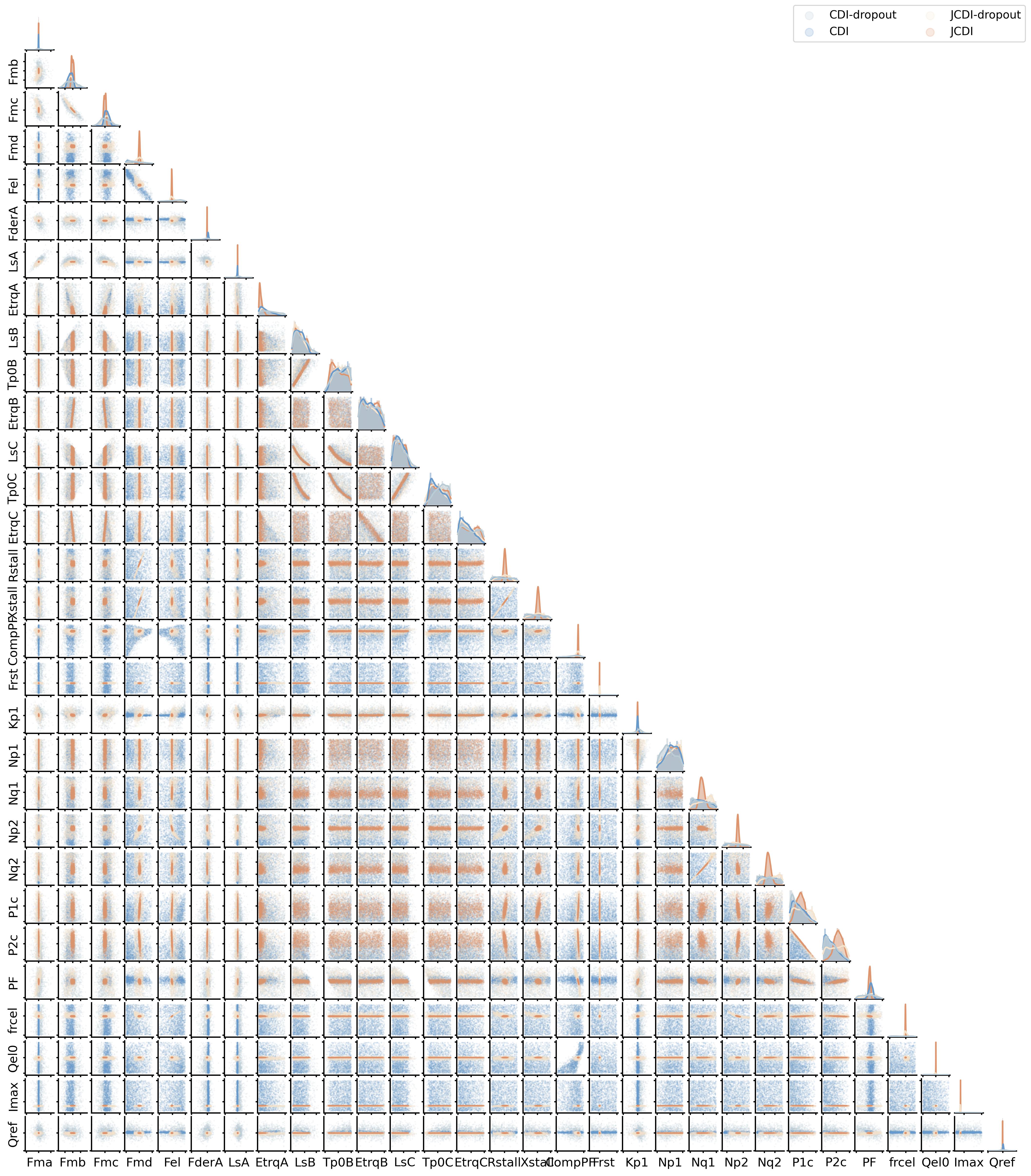} 
	\caption{\textbf{Posterior results of parameter estimation considering 10\% data dropout.} 
    Parameters estimates of CDI and JCDI without data dropout are shown in blue and orange, respectively, while the respective estimation results of CDI and JCDI considering 10\% data dropout are presented in green and yellow. \textcolor {cyan}{The statistical plots are constructed using 1000 parameter estimates.}}
	\label{fig:parameter-correlation-dropout}
\end{figure}

\begin{figure}[htbp]
	\centering
    \includegraphics[width=0.95\textwidth]{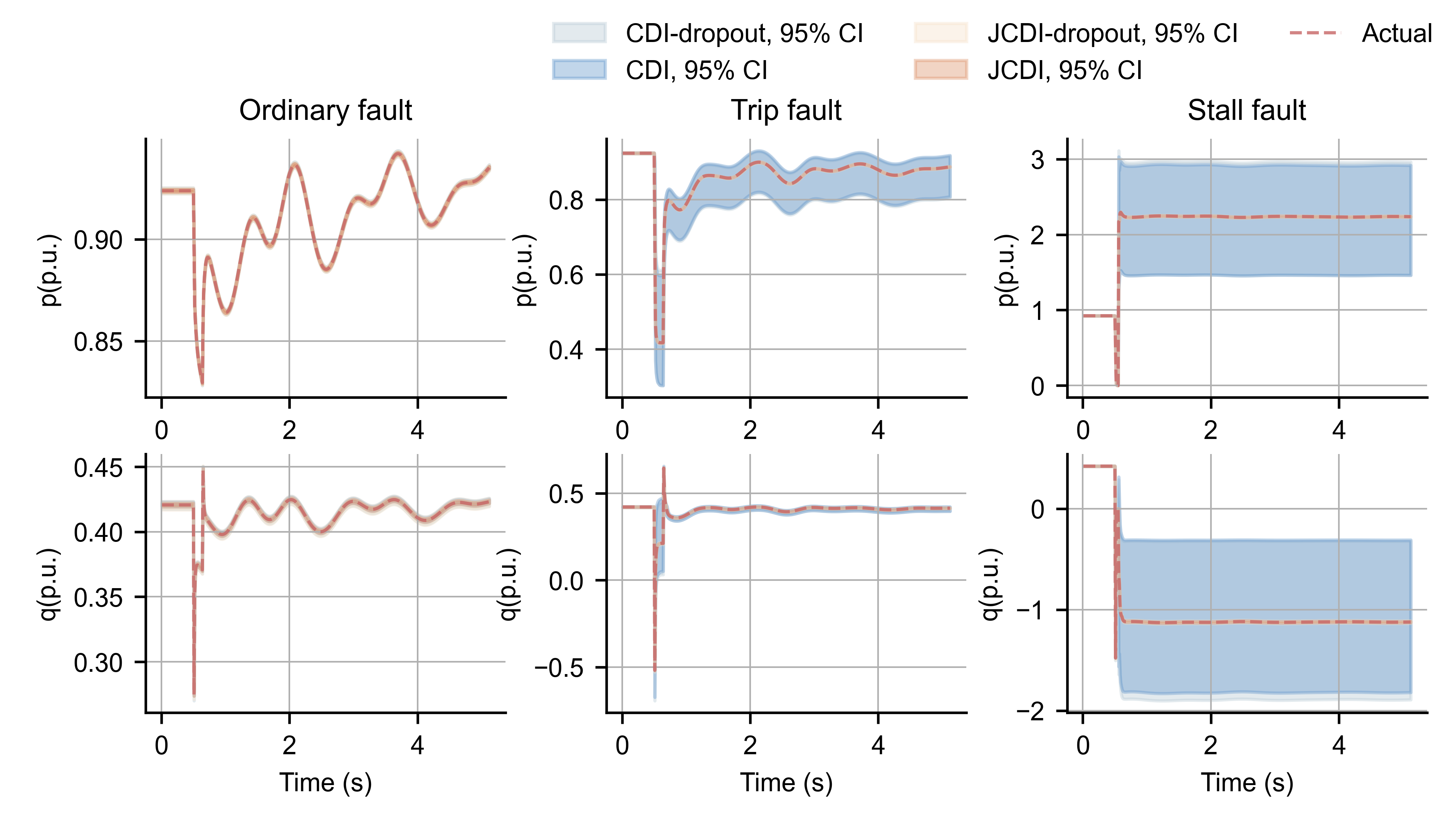} 
\caption{\textbf{Prediction results of dynamic responses considering 10\% data dropout.} 
    The pink dashed lines are the actual power measurements. When inferred with the complete measurements, the predicted power trajectories of CDI and JCDI with 95\% \textcolor {cyan}{credible intervals (CIs)} are displayed with blue and orange areas, respectively. The respective prediction results of CDI and JCDI with 10\% data dropout are presented by green and yellow areas. \textcolor {cyan}{The plots are constructed using 1000 data samples.}}
	\label{fig:traj-result-dropout}
\end{figure}

\bmhead{\textcolor {cyan}{Impact of data dropout}} 
We construct synthetic data with dropout by randomly removing 10\% of the data points in the measurements and expanding the training dataset size to 480000 to avoid overfitting.
Supplementary Figures~\ref{fig:parameter-correlation-dropout} and~\ref{fig:traj-result-dropout} respectively show the complete posterior results of parameter estimation and prediction results of dynamic responses considering 10\% data dropout. Supplementary Table~\ref{tab:cmp-dropout} summarizes the estimation accuracy.
\begin{table}[htbp]
	\caption{Evaluation of estimation performance with 10\% data dropout}
	\centering
	\begin{tabular}{lllll}
		\toprule
		\multirow{2}{*}{\makecell[l]{Parameteri-\\ zation methods}} & \multirow{2}{*}{\makecell[l]{MARPEs of\\ parameters}} &\multicolumn{3}{c}{RMSEs of power trajectories}   \\
		\cmidrule(r){3-5}
	  & & Ordinary fault & Trip fault  &  Stall fault \\
		\midrule
        CDI        &  18.0\%  & $1.01 \times {10^{ - 3}}$    & $4.44 \times {10^{ - 2}}$    &    $7.97 \times {10^{ - 1}}$   \\
        CDI-dropout    &  20.1\%  &   $1.76 \times {10^{ - 3}}$    &  $4.63 \times {10^{ - 2}}$      &  $7.39 \times {10^{ - 1}}$  \\ \hline
	JCDI         & 7.46\%   &  $7.84 \times {10^{ - 4}}$    &  $8.61 \times {10^{ - 4}}$    &    $3.75 \times {10^{ - 3}}$   \\
	JCDI-dropout     & 13.6\%   & $1.90 \times {10^{ - 3}}$    &  $2.12 \times {10^{ - 3}}$   &      $8.21 \times {10^{ - 3}}$  \\
		\bottomrule
	\end{tabular}
	\label{tab:cmp-dropout}
\end{table}
\section{Supplementary information on comparative studies}
\label{sec:Appendix-D}

\subsection{Supplementary information on comparative methods}
\label{sec:supp-cmp-alg}

\bmhead{Reinforcement learning}
We formulate the parameter calibration process for CMPLDWG as a Markov decision process. The agent starts from an initial parameter estimation state, modifying the parameters at each step. Then, it transfers to the next state of estimation and obtains a reward signal in this transition. Through this process, the agent learns a policy capable of adjusting model parameters to minimize observation error. The state is defined as the current estimation of parameters as expressed by~\eqref{s}.

\begin{equation}
s = \hat \theta \label{s}.
\end{equation}
Action is defined as the adjustment of parameters, which is expressed by~\eqref{a}.
\begin{equation}
a = \Delta \hat \theta  \label{a}.
\end{equation}
State transition represents the parameter update as shown in~\eqref{transition}.
\begin{equation}
s' = \hat \theta ' = \hat \theta  + \Delta \hat \theta   \label{transition}.
\end{equation}
Reward is calculated based on the change in RMSE of dynamic power responses (formulated in equation~(\ref*{RMSE})). To motivate continuous accuracy improvement when RMSE is small, a reciprocal representation is used as shown in~\eqref{r}.
\begin{equation}
r =  - \Delta \frac{{1}}{{RMSE{\rm{ + }}0.1}}   \label{r}.
\end{equation}

We use Deep Q-Network (DQN), a deep reinforcement learning algorithm, to solve the parameterization problem~\cite{DQN}. Q-learning is a value-based reinforcement learning method that updates the action-value function (Q-function) using the temporal-difference learning approach as expressed by~\eqref{Q-update}. DQN uses a neural network (Q-network) to approximate the Q-function and trains the Q-network based on the temporal-difference error. In this work, the Q-function is approximated by a fully connected neural network, including two hidden layers with 512 and 256 units, respectively. The $\varepsilon$-greedy policy with a piecewise exploration schedule is used to realize the trade-off between exploration and exploitation. The training parameters for DQN are listed in Supplementary Table~\ref{tab:hyper-DQN}.
When DQN is trained under multi-fault event condition (MEC), the fault events are selected randomly for each episode.
\begin{equation}
Q\left( {s,a} \right) \leftarrow Q\left( {s,a} \right) + \alpha  \cdot \left( {r + \gamma  \cdot {{\max }_a}Q\left( {s',a} \right)\left. { - Q\left( {s,a} \right)} \right)} \right.\label{Q-update},
\end{equation} 
\noindent where $Q\left( {s,a} \right)$ denotes the action-value function, $\alpha $ represents the learning rate, and $\gamma$ is the discount factor.

\begin{table}[htbp]
	\caption{Hyperparameters for DQN}
	\centering
	\begin{tabular}{llll}
		\toprule
		Hyperparameters     & Notations         & Values \\
		\midrule
        Batch size     &  $B$      &  32 \\
		\multirow{2}{*}{Learning rate}     &  \multirow{2}{*}{$\alpha $}    & \multirow{2}{*}{\makecell[l]{Decay from $ 1 \times {10^{ - 3}} $\\ to $5 \times {10^{ - 5}}$}  }    \\\\
        Replay buffer size     &   ${B_r}$     &  5000 \\
        Target update frequency &  ${f_T}$      &  500 \\
        \multirow{2}{*}{Exploration}  &  \multirow{2}{*}{$\varepsilon$}  & \multirow{2}{*} {\makecell[l]{Decay from 1\\ to 0.01}  }    \\\\
        Discount rate & $\gamma$  & 0.99 \\
		\bottomrule
	\end{tabular}
	\label{tab:hyper-DQN}
\end{table}

\begin{figure}[htbp]
	\centering
    \includegraphics[width=10cm]{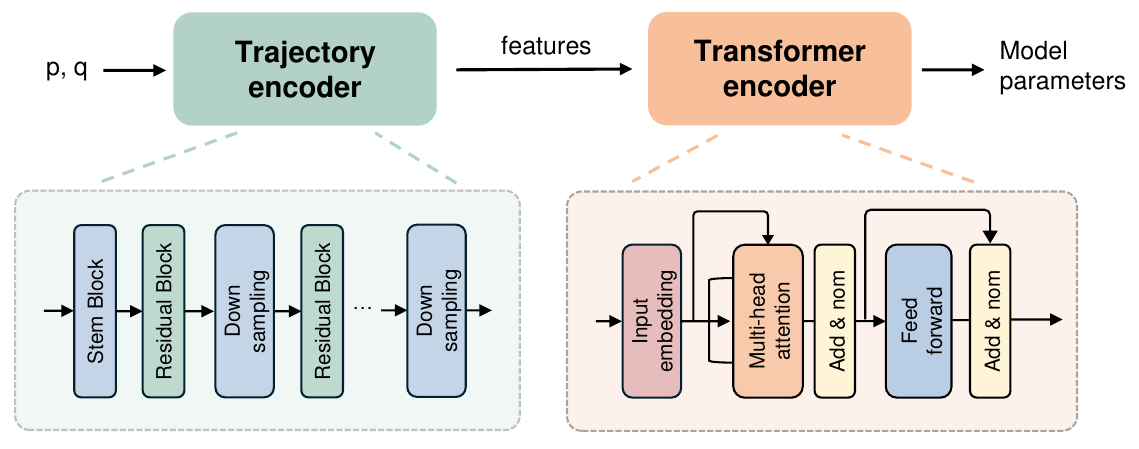} 
	\caption{\textbf{The neural network structure for SL.} The neural network for SL, Res-TFR, consists of a ResNet-based trajectory encoder in series with a transformer encoder. It learns the mapping between power trajectories and model parameters.}
	\label{fig:structure-SL}
\end{figure}

\bmhead{Supervised learning}
Supplementary Figure~\ref{fig:structure-SL} illustrates the neural network structure for supervised learning (SL). A trajectory encoder based on a Residual Network in series with a transformer encoder (Res-TFR) extracts features from active and reactive power trajectories. These features are then tokenized and processed by a transformer encoder to infer the model parameters. Supplementary Table~\ref{tab:hyper-SL} lists the algorithm parameters.

\begin{table}[htbp]
	\caption{Hyperparameters for Res-TFR}
	\centering
	\begin{tabular}{lllll}
		\toprule
		Parameter types&Hyperparameters     & Notations         & Values \\
		\midrule
        \multirow{2}{*}{Training} &Batch size     &  $B$      &  128  \\
        &Learning rate              &  $lr$      &  $1 \times {10^{ - 4}}$  \\
        \midrule
        \multirow{3}{*}{\makecell[l]{Trajectory\\ encoder}} &  Stem layer (kernel size, number)    &  -     &  [$1\times4$, 32]  \\
        &Downsampling layer 1    &  -     &  [$1\times4$, 128]  \\
        &Downsampling layer 2    &  -     &  [$1\times4$, 256]  \\
        \midrule
        \multirow{4}{*}{\makecell[l]{Transformer\\ encoder}} &Number of layers   &  -     &  3  \\
        &Number of heads  &  -     &  4  \\
        &Features in the input   &  -     &  256  \\
        &Features in the feed-forward network   &  -     &  512  \\
		\bottomrule
	\end{tabular}
	\label{tab:hyper-SL}
\end{table}

\bmhead{Flow-Matching Posterior Estimation (FMPE)} 
FMPE uses continuous normalizing flow as the conditional density estimator to estimate the parameter posterior from observations~\cite{NEURIPS2023-FMPE2}.  
It learns a vector field that transforms a simple base distribution into the target data distribution by minimizing the flow matching loss, which enables faster training.
The implementation of FMPE for our CMPLDWG parameterization problem relies on the Simulation-Based Inference (sbi) package~\cite{BoeltsDeistler_sbi_2025}.
We adopt the default multilayer perceptron neural network structure with the hyperparameters listed in Supplementary Table~\ref{tab:hyper-FMPE}. 

\begin{table}[htbp]
	\caption{Hyperparameters for FMPE}
	\centering
	\begin{tabular}{lllll}
		\toprule
		Parameter types&Hyperparameters     & Notations         & Values \\
		\midrule
        \multirow{2}{*}{Training} &Batch size     &  $B$      &  128  \\
        &Learning rate              &  $lr$      &  $1 \times {10^{ - 4}}$  \\
        \midrule
        \multirow{2}{*}{\makecell[l]{Neural network\\ configuration}} &  Number of layers    &  -     &  5  \\
        &Number of hidden features    &  -     &  256  \\
		\bottomrule
	\end{tabular}
	\label{tab:hyper-FMPE}
\end{table}

\bmhead{Approximate Bayesian Computation-Sequential Monte Carlo (ABC-SMC)}
ABC-SMC is a likelihood-free Bayesian inference method that infers posterior distributions with sequential Monte Carlo sampling~\citep{ABC-SMC}.
Parameter estimation with ABC-SMC is also implemented based on the sbi package with the suggested parameters in the benchmark (Supplementary Table~\ref{tab:hyper-ABC-SMC})~\cite{BoeltsDeistler_sbi_2025, lueckmann2021benchmarkingsimulationbasedinference}.
We use ${l_2}{\rm{ - norm}}$ as the metric to measure the distance between simulated and actual dynamic responses and Gaussian kernel to perturb intermediate samples. When ABC-SMC is trained under MEC, the distances under different fault events are averaged.

\begin{table}[htbp]
	\caption{Hyperparameters for ABC-SMC}
	\centering
	\begin{tabular}{llll}
		\toprule
		Hyperparameters     & Notations         & Values \\
		\midrule
        Population size     &  -      &  100 \\
        Epsilon decay     &  -      &  0.2 \\
        Kernel variance scale     &  -      &  0.5 \\
        Number of simulations     &  -      &  10000 \\
		\bottomrule
	\end{tabular}
	\label{tab:hyper-ABC-SMC}
\end{table}

\bmhead{Covariance matrix adaptation evolution strategy (CMA-ES)}
CMA-ES is an evolutionary algorithm that iteratively solves the optimization problem by adaptively updating the covariance matrix of the multi-variable distributions~\citep{hansen2006eda}.
The objective function is calculated by the RMSE of power trajectories under single-fault event condition (SEC), while the mean RMSE for different fault events when it is trained under MEC.
We use the default parameters of the algorithm. 
The population size is calculated by $4 + floor\left( {3 \cdot \ln \left( N \right)} \right)$, where $N$ is the number of variables and equals 30 in our problem.

\subsection{Supplementary results}
\label{sec:supp-cmp-results}

\bmhead{Training progress} 
JCDI is trained for 4000 epochs until the training and testing losses level off, which costs nearly four days on a RTX 3090 GPU. To implement a fair comparison, Res-TFR and FMPE are also trained for 4000 epochs. They are categorized as amortized methods, which are trained on a large dataset of parameter-observation pairs, and can estimate parameters from different observations without the need for retraining.
However, DQN, ABC-SMC, and CMA-ES are sequential methods that require iterative simulations during the training process and focus on parameter inference for an individual observation. We run them for around four days (on CPU) under MEC, matching the computation time of JCDI.
Their performance evolutions during the training process are shown in Supplementary Figure~\ref{fig:evolution-algs}. The training and testing losses for FMPE and Res-TFR also stabilize after 4000 epochs. There is no further noticeable improvement in the objective functions for CMA-ES and DQN. ABC-SMC does not converge well. The mean trajectory RMSE reduces with an increase of simulation numbers before 10000 and becomes larger afterwards. Therefore, the result under 10000 simulations is used for comparison.

\begin{figure}[htbp]
	\centering
    \includegraphics[width=0.95\textwidth]{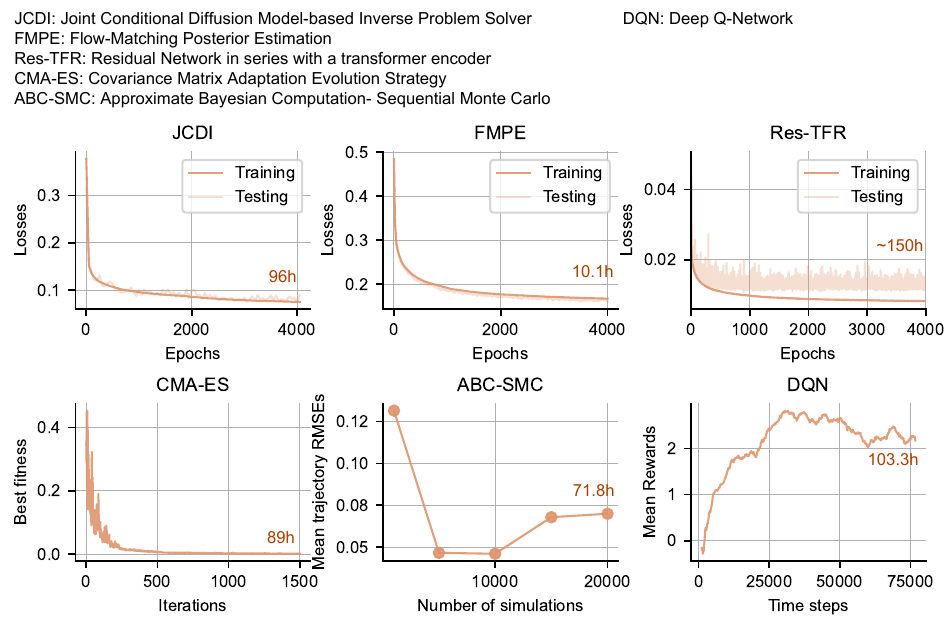} 
	\caption{
    \textbf{The evolution process of different parameterization methods.} 
    }
	\label{fig:evolution-algs}
\end{figure}

\begin{figure}[htbp]
	\centering
     \includegraphics[width=0.98\textwidth]{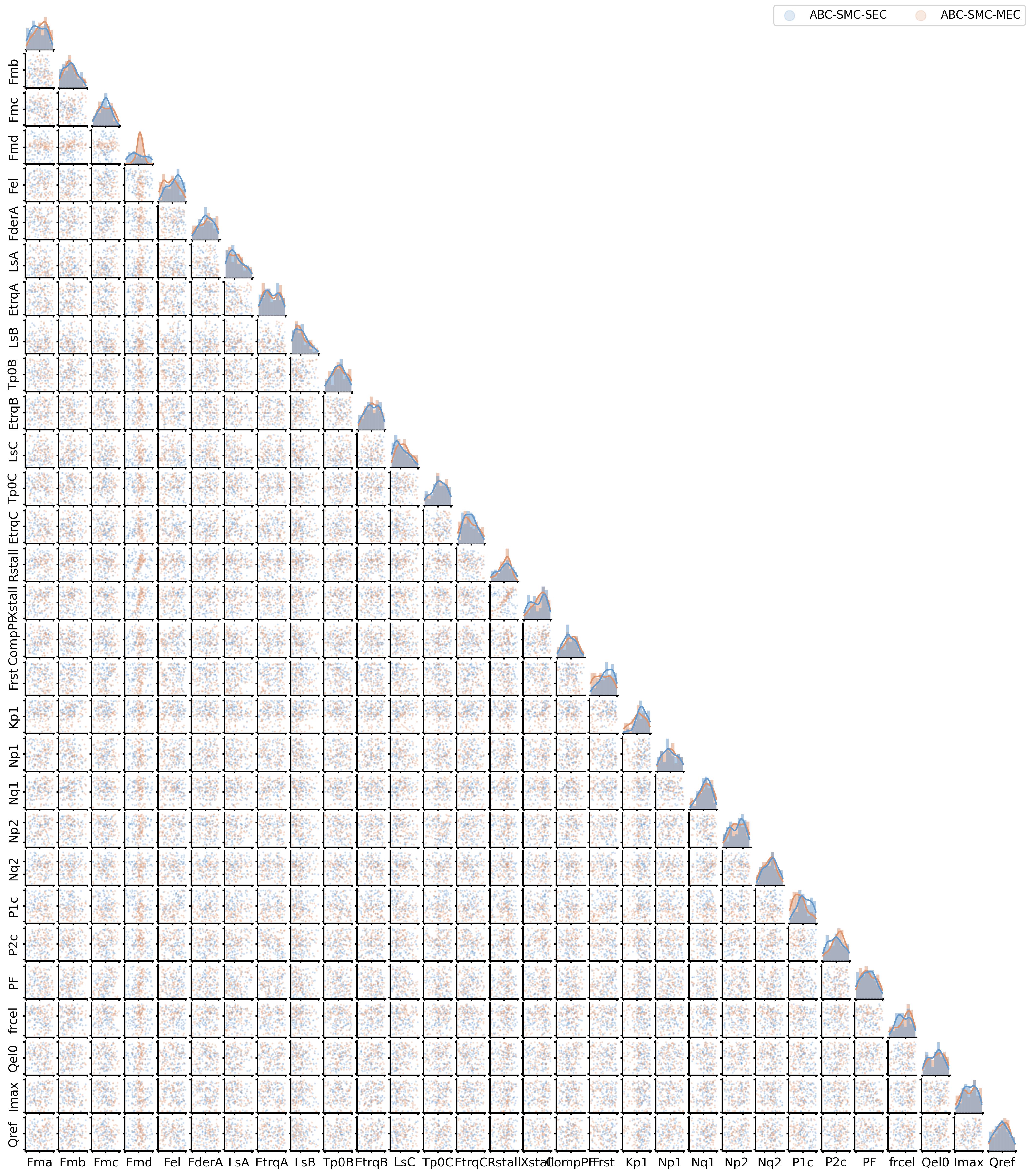} 
	\caption{\textbf{Posterior results of parameter estimation for ABC-SMC.}
    \textcolor {cyan}{Parameters estimates of ABC-SMC under single-fault event condition (SEC) and multi-fault event condition (MEC) are shown in blue and orange, respectively. The statistical plots are constructed using 100 data samples.
    }}
	\label{fig:parameter-correlation-SMCABC}
\end{figure}

\begin{figure}[htbp]
	\centering
     \includegraphics[width=0.98\textwidth]{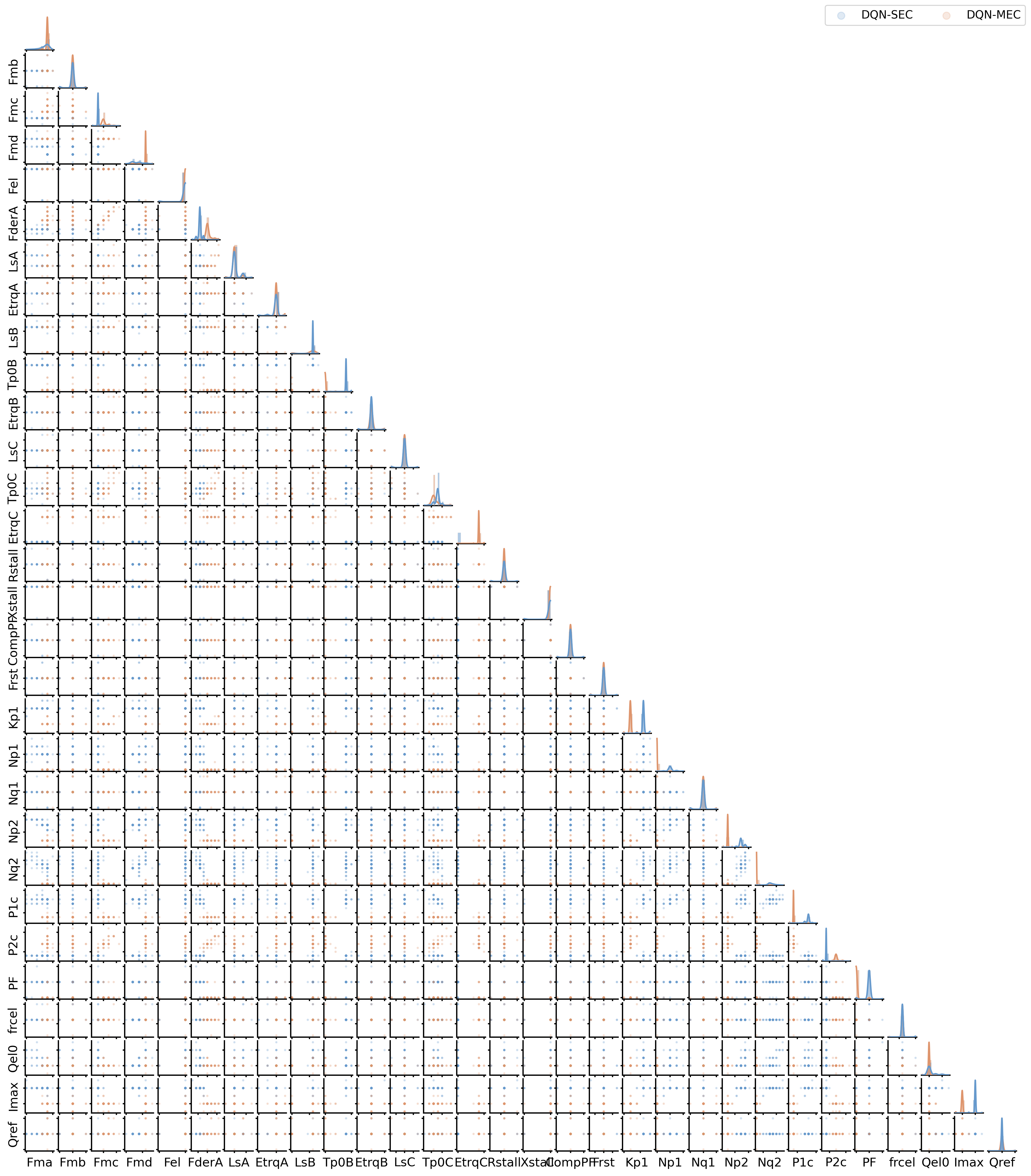} 
	\caption{\textbf{Posterior results of parameter estimation for DQN.} 
   \textcolor {cyan}{Parameters estimates of DQN under single-fault event condition (SEC) and multi-fault event condition (MEC) are shown in blue and orange, respectively. The statistical plots are constructed using 100 data samples.
    }}
	\label{fig:parameter-correlation-DQN}
\end{figure}

\bmhead{Parameter estimation uncertainties} 
The representative parameter posterior results across different parameterization methods are compared in Supplementary Figure~\ref{fig:correlation-cmps}.
The complete parameter posterior results of FMPE, ABC-SMC, and DQN are presented in Supplementary Figures~\ref{fig:parameter-correlation-FMPE},~\ref{fig:parameter-correlation-SMCABC} and~\ref{fig:parameter-correlation-DQN}, respectively. 

{
\color{cyan}
In keeping with previous studies, we generate 1000 parameter estimates for FMPE. 
For ABC-SMC, the final population, with a population size of 100, is used for analysis.
Due to high computational burdern for DQN, 100 solutions are deduced by rerunning the well-trained DQN agent.
While notable associations among parameter pairs ${F_{{\rm{mb}}}}$ and ${F_{{\rm{mc}}}}$, ${F_{{\rm{md}}}}$ and ${F_{{\rm{el}}}}$, $CompPF$ and ${Q_{{\rm{el0}}}}$, and ${F_{{\rm{mb}}}}$ and ${L_{{\rm{sA}}}}$ estimated by CDI were observed in previous analysis, we note these associations weakened because of the increase of parameter uncertainties for FMPE conditioned on a single fault event, where the association between ${F_{{\rm{mb}}}}$ and ${L_{{\rm{sA}}}}$ becomes inapparent.
ABC-SMC produces parameter estimates with higher uncertainties, while parameters estimated by DQN exhibit a large deviation from the actual values -- both failing to reveal these relations.
When comparing the results under single- and multi-fault event conditions, we demonstrate that JCDI effectively mitigates parameter non-identifiability and yields more accurate parameter estimation than CDI.
The effect of MEC also holds for FMPE: the posterior distributions of ${F_{{\rm{md}}}}$, ${F_{{\rm{el}}}}$, $CompPF$, and ${Q_{{\rm{el0}}}}$ become much more concentrated than those estimated using a single-fault event. However, FMPE still fails to accurately identify parameters such as ${F_{{\rm{mb}}}}$, ${F_{{\rm{mc}}}}$, and ${L_{{\rm{sA}}}}$.
For ABC-SMC, MEC helps reduce the estimation uncertainties of parameters, including ${F_{{\rm{md}}}}$, ${R_{{\rm{stall}}}}$, and ${X_{{\rm{stall}}}}$, which present high sensitivity under the stall fault, but it fails to work for other parameters.
Large parameter deviations are observed for DQN under both SEC and MEC, while ${F_{{\rm{md}}}}$ gets closer to its actual value when DQN is trained with multiple faults.

For further evaluation of the calibration performance of different methods, we present inference results for all model parameters under MEC in Supplementary Figure~\ref{fig:posterior-cmps}. 
JCDI produces more concentrated distributions around the actual values than other probabilistic methods for most of the parameters.
Notably, the posterior distributions for parameters such as ${F_{{\rm{mb}}}}$ and ${E_{{\rm{trqB}}}}$ show multiple peaks, which can be caused by the parameter dependencies and JCDI's probabilistic nature.
FMPE and ABC-SMC present high estimation uncertainties for parameters such as ${F_{{\rm{mb}}}}$, ${F_{{\rm{mc}}}}$ and $PF$. 
While failing to estimate a number of parameters, DQN implements accurate estimations for ${L_{{\rm{sC}}}}$ and $PF$.
Some parameter estimates for CMA-ES and Res-TFR are close to the actual values, such as ${F_{{\rm{mb}}}}$, ${F_{{\rm{mc}}}}$, ${F_{{\rm{rst}}}}$, ${Q_{{\rm{el0}}}}$, and ${I_{{\rm{max}}}}$.
Nevertheless, obvious deviations are also observed for parameters estimated by CMA-ES, such as ${F_{{\rm{md}}}}$, ${F_{{\rm{el}}}}$, ${L_{{\rm{sA}}}}$, ${R_{{\rm{stall}}}}$, and ${X_{{\rm{stall}}}}$ -- some of which are dependent as previously demonstrated. 
Res-TFR maintains high inference accuracy for these parameters, but it exhibits large estimation errors for ${P_{{\rm{1c}}}}$ and ${P_{{\rm{2c}}}}$.
}
\begin{figure}[htbp]
	\centering
    \includegraphics[width=0.95\textwidth]{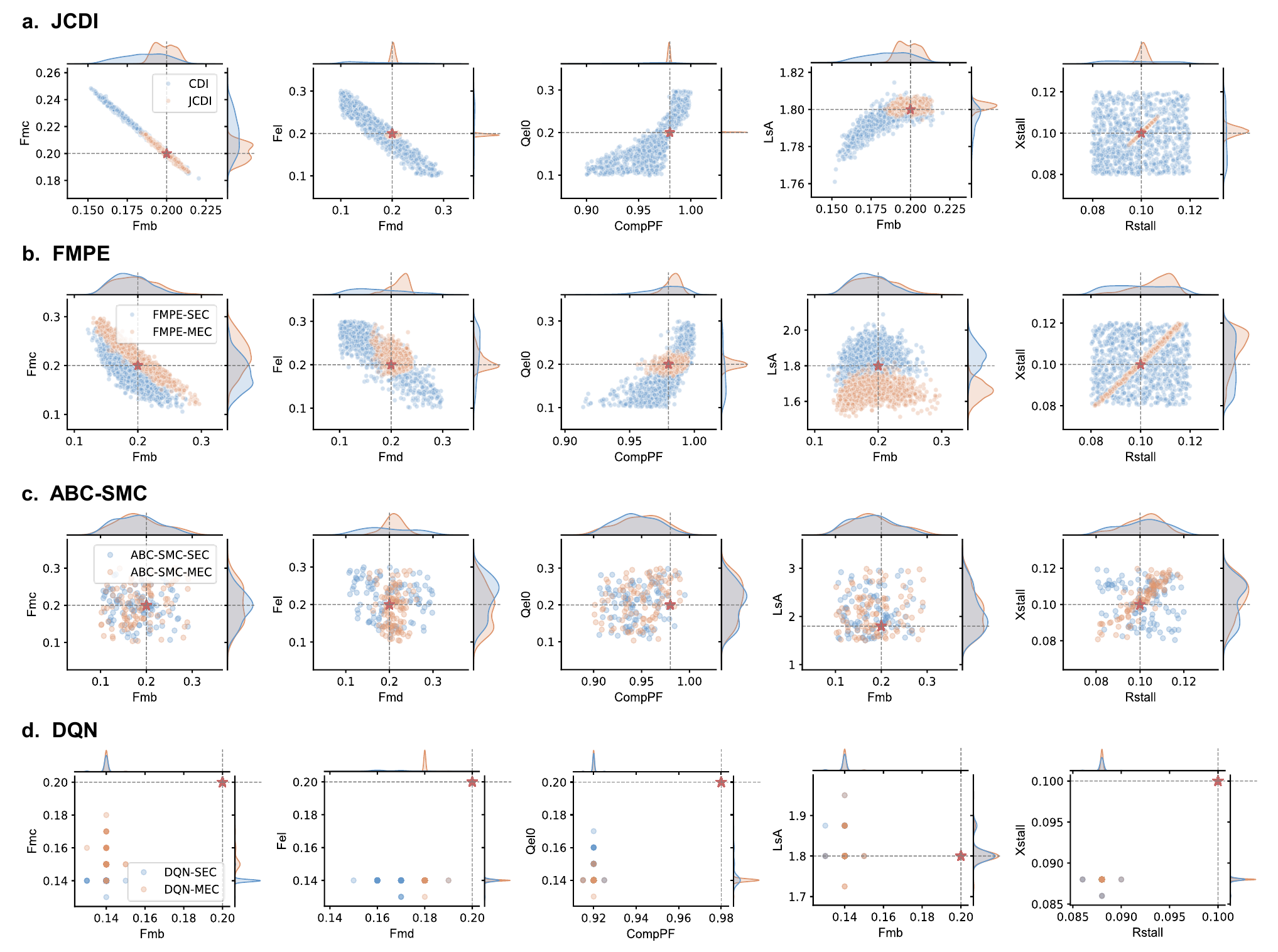} 
	\caption{\textbf{Comparison of representative parameter posterior results across different parameterization methods.} 
    \textcolor {cyan}{Scatter plots with marginal distributions} of selected parameter pairs are compared for different parameterization methods, including JCDI, FMPE, ABC-SMC, and DQN.
    Blue and orange dots represent parameters estimated under \textcolor {cyan}{single-fault event condition (SEC) and multi-fault event condition (MEC)} for different methods, respectively. 
    The pink stars indicate the actual values of model parameter pairs.
    \textcolor {cyan}{The statistical figures for JCDI and FMPE are constructed using 1000 data samples, while they are plotted with 100 samples of parameter estimates for ABC-SMC and DQN.}
    }
	\label{fig:correlation-cmps}
\end{figure}

\begin{figure}[htbp]
	\centering
     \includegraphics[width=0.98\textwidth]{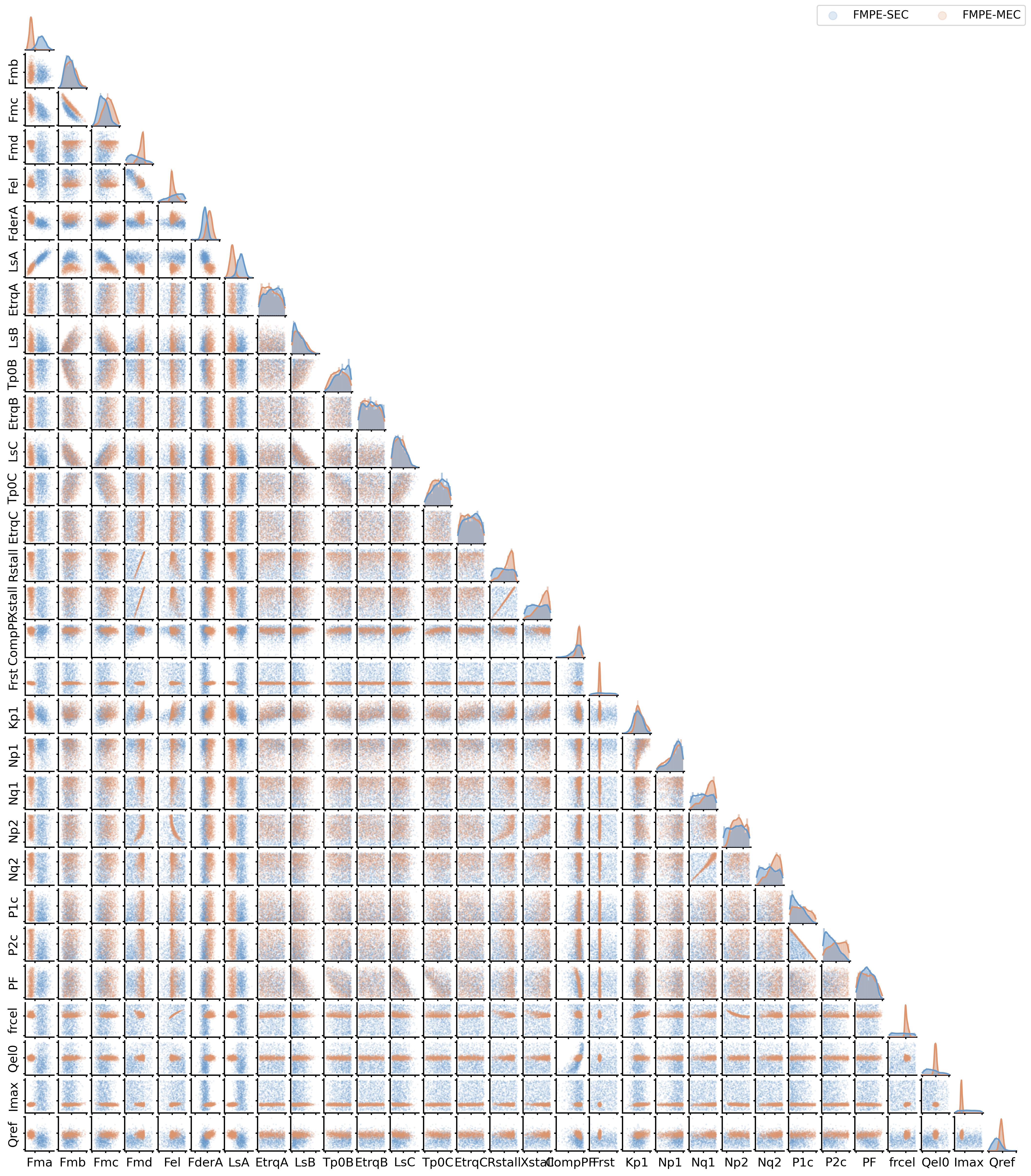} 
	\caption{\textbf{Posterior results of parameter estimation for FMPE.} 
   \textcolor {cyan}{Parameters estimates of FMPE under single-fault event condition (SEC) and multi-fault event condition (MEC) are shown in blue and orange, respectively. The statistical plots are constructed using 1000 data samples.}}
	\label{fig:parameter-correlation-FMPE}
\end{figure}

\begin{figure}[!htp]
	\centering
    \includegraphics[width=0.95\textwidth]{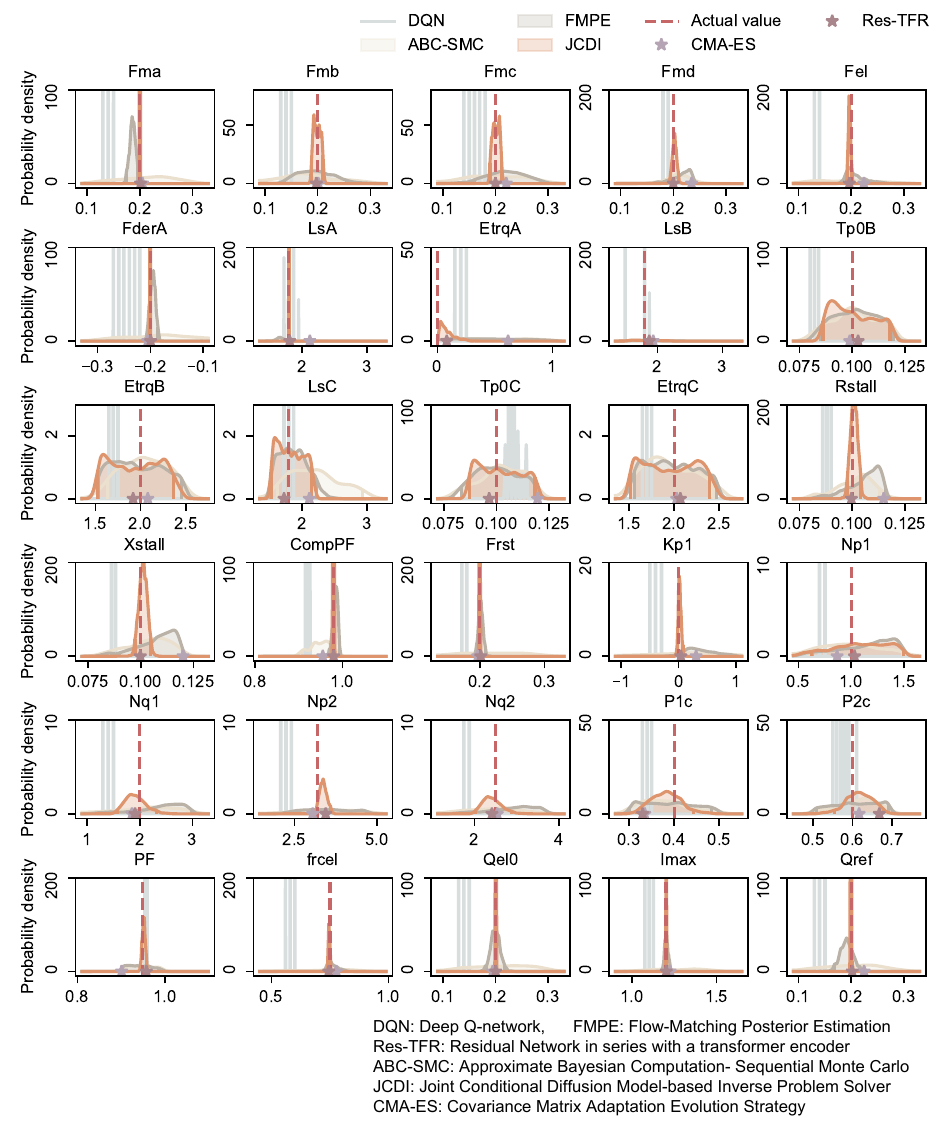} 
	\caption{\textcolor {cyan}{
    \textbf{Comparison of the complete parameter inference results across different parameterization methods under multi-fault event condition.}} 
    Light gray, brown, yellow, and orange curves respectively represent the parameter posterior distributions under DQN, FMPE, ABC-SMC, and JCDI with the corresponding colored areas indicating 95\% credible intervals. 
    The parameter estimates of Res-TFR and CMA-ES are denoted by dark brown and purple stars, respectively. 
    The actual model parameters values are displayed as pink dashed lines. 
    \textcolor {cyan}{1000 data samples are used for FMPE and JCDI, while the parameter posterior distributions are plotted with 100 parameter estimates for DQN and ABC-SMC.}
    }
	\label{fig:posterior-cmps}
\end{figure}

\bmhead{Dynamic response prediction} 
The dynamic responses predicted by FMPE, ABC-SMC, DQN, Res-TFR, and CMA-ES are displayed in Supplementary Figure~\ref{fig:traj-result-cmp-algs}a-e.
\textcolor {cyan}{When trained under MEC, the power trajectories predicted by JCDI, FMPE, and CMA-ES are all close to the actual trajectories under different fault disturbances.
However, ABC-SMC presents wider credible intervals of predicted power trajectories, while DQN shows evident trajectory deviation.}

\begin{figure}[htbp]
	\centering
    \includegraphics[width=0.95\textwidth]{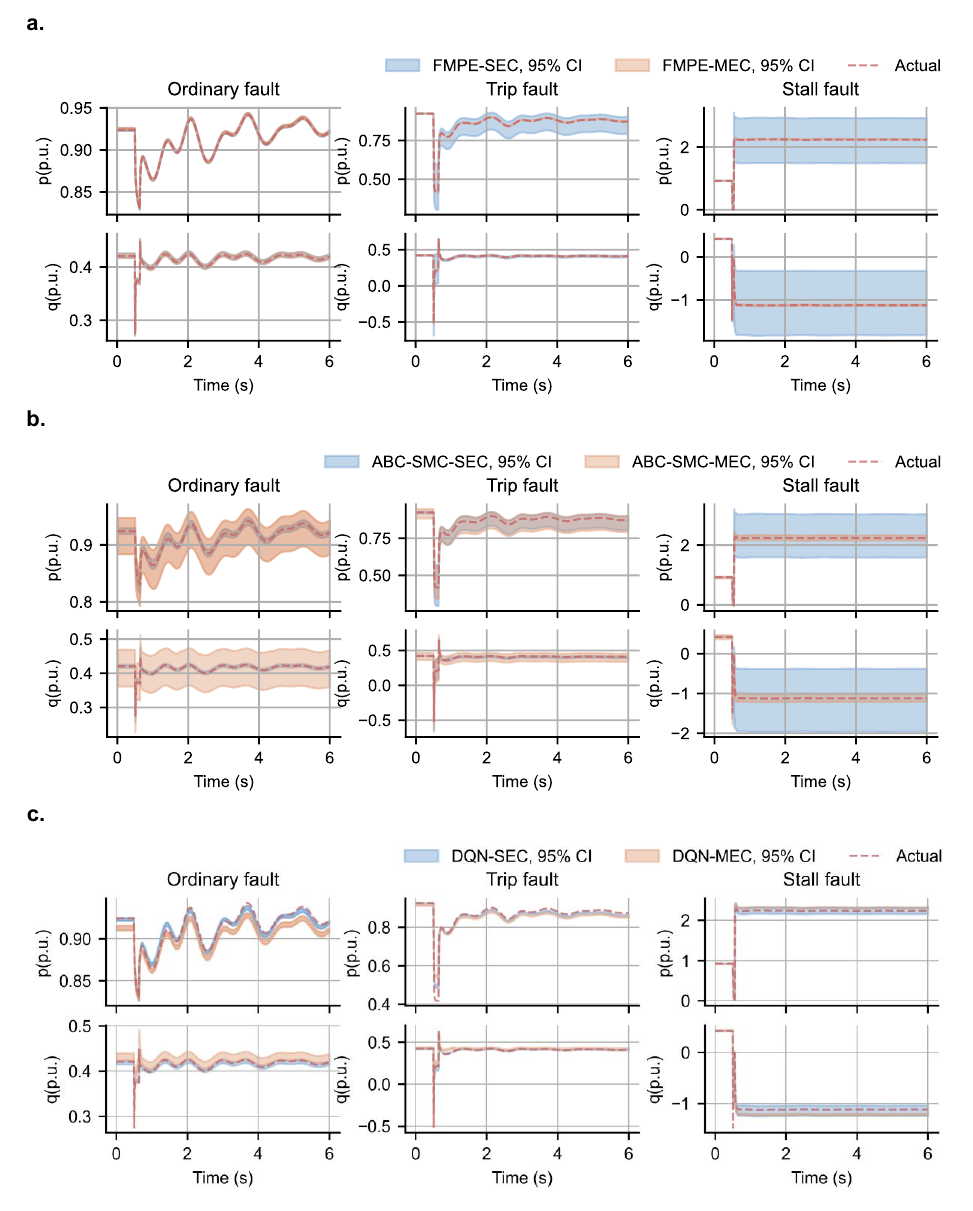} 
\caption{\textbf{Prediction results of dynamic responses across different parameterization methods.} 
    \textcolor {cyan}{\textbf{a-e} respectively show the prediction results of FMPE, ABC-SMC, DQN, Res-TFR and CMA-ES. The pink dashed lines represent the actual power measurements. In \textbf{a-c}, the predicted power trajectories for different parameterization methods under single-fault event condition (SEC) and multi-fault event condition (MEC) with 95\% credible intervals (CIs) are displayed with blue and orange areas, respectively. \textbf{a} is constructed using 1000 data samples, \textbf{b} and \textbf{c} are constructed using 100 data samples.
    }}
	\label{fig:traj-result-cmp-algs}
\end{figure}
\begin{figure}[h]
	\ContinuedFloat
	\centering
    \includegraphics[width=0.95\textwidth]{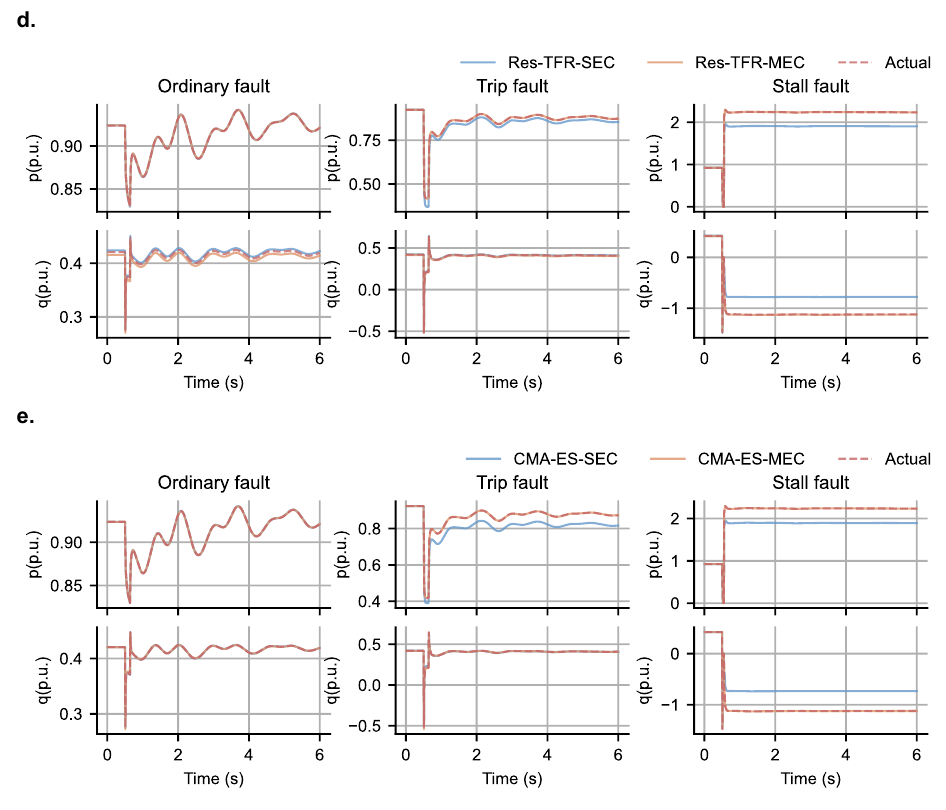} 
\caption{\textbf{Prediction results of dynamic responses across different parameterization methods.} 
    \textcolor {cyan}{In \textbf{d-e}, the predicted power trajectories under SEC and MEC are respectively represented with blue and orange lines.
    }}
	\label{fig:traj-result-cmp-algs}
\end{figure}


\clearpage
\newpage
\bibliography{references}